\documentclass{article}
\usepackage{adjustbox}
\PassOptionsToPackage{numbers, compress}{natbib}
\usepackage[main,preprint]{neurips_2026}
\usepackage{amssymb}
\usepackage[utf8]{inputenc} % allow utf-8 input
\usepackage[T1]{fontenc}    % use 8-bit T1 fonts
\usepackage{hyperref}       % hyperlinks
\usepackage{url}            % simple URL typesetting
\usepackage{booktabs}       % professional-quality tables
\usepackage{amsfonts}       % blackboard math symbols
\usepackage{nicefrac}       % compact symbols for 1/2, etc.
\usepackage{microtype}      % microtypography
\usepackage{xcolor}         % colors
\usepackage{amsmath}
\usepackage{stmaryrd}
\usepackage{algorithm}
\usepackage{algpseudocode}
 \usepackage{multirow}

 \usepackage[table]{xcolor}
\usepackage{pifont} % optional, if you want symbol alternatives
\usepackage[T1]{fontenc}
\usepackage[utf8]{inputenc}
\usepackage{listings}
\usepackage{caption}
\lstdefinestyle{goalmemprompt}{
  basicstyle=\ttfamily\scriptsize,
  breaklines=true,
  breakatwhitespace=false,
  columns=fullflexible,
  keepspaces=true,
  showstringspaces=false,
  frame=single,
  literate={—}{{---}}1
}
\usepackage{booktabs,tabularx,hyperref}
\title{Goal-Oriented Reasoning for RAG-based Memory in Conversational Agentic LLM Systems}

\author{%
Jiazhou Liang\thanks{Equal contribution.} \\
University of Toronto \\
\texttt{joe.liang@mail.utoronto.ca}
\And
Armin Toroghi\footnotemark[1] \\
University of Toronto \\
\texttt{armin.toroghi@mail.utoronto.ca}
\And
Yifan Simon Liu \\
University of Toronto \\
\And
Faeze Moradi Kalarde \\
University of Toronto \\
\And
Liam Gallagher \\
University of Toronto \\
\And
Scott Sanner \\
University of Toronto \\
Vector Institute for Artificial Intelligence
}

\usepackage[most]{tcolorbox}

\newtcolorbox{flowexample}{%
  enhanced,
  breakable,
  colback=black!2,
  colframe=black!35,
  boxrule=0.4pt,
  arc=0.5mm,
  left=1.5mm,
  right=1.5mm,
  top=1mm,
  bottom=1mm,
  before skip=2pt,
  after skip=2pt,
  fontupper=\small,
}
\begin{document}

\maketitle

\begin{abstract}
LLM-based conversational AI agents struggle to maintain coherent 
%and personalized 
behavior 
over long horizons due to limited context. While RAG-based approaches are increasingly adopted to overcome this limitation by storing interactions in external memory modules and performing retrieval from them,  their effectiveness in answering challenging questions (e.g., multi-hop, commonsense, etc.) ultimately depends on the agent’s ability to reason over the retrieved information. 
However, existing methods typically retrieve memory by semantic similarity to the raw user utterance, which lacks explicit reasoning over missing intermediate facts and often returns evidence that is irrelevant or insufficient for grounded reasoning.
% However, existing works largely rely on raw user utterances driven semantic similarity for retrieval which lacks the ability to identify missing intermediate facts, and often retrieves irrelevant or insufficient evidence for grounded reasoning. 
In this work, we introduce \textbf{\textsc{Goal-Mem}}, a goal-oriented reasoning framework for RAG-based agentic memory that performs explicit backward chaining from the user’s utterance as a goal. Rather than progressively expanding from retrieved context, \textsc{Goal-Mem} decomposes each goal into atomic subgoals, performs targeted memory retrieval to satisfy each subgoal, and iteratively identifies what information from memory should be retrieved when intermediate goals cannot be resolved. We formalize this process in Natural Language Logic, a logical system that combines the verifiability of reasoning provided by FOL with the expressivity of natural language.
%We formalize this process in an extended natural language logic (NL-Logic), where goals and memory are represented as typed natural language terms, enabling verifiable reasoning, constraining grounding through type consistency, and incorporating commonsense axioms from the LLM to bridge gaps between stored memory and required knowledge. \textsc{Goal-Mem} is plug-and-play with existing RAG-based memory backbones and requires no modification to their storage mechanisms. 
Through extensive experiments on two datasets, comparing to eight strong memory baselines, we show that \textsc{Goal-Mem} consistently improves performance, particularly on tasks requiring multi-hop reasoning and implicit inference.
%Our results demonstrate that backward, goal-directed retrieval provides a more effective and robust paradigm for memory-grounded reasoning in agentic LLM systems.
 % Existing agentic memory methods rely on forward chaining -style reasoning, which indiscriminately retrieves facts and often expands the search space unnecessarily. We take a goal-directed backward chaining approach that explicitly identifies the minimal set of missing facts required to ground a query, generates commonsense axioms to compensate for incompleteness of the memory, and constructs proper subgoals when the current goal cannot be satisfied to perform backward chaining.
\end{abstract}

\section{Introduction}
Large Language Models (LLMs) are increasingly deployed as conversational and agentic systems for complex, long-horizon tasks~\citep{maharana2024evaluating,yao2025taubench,guo2025vogue,liu2026semantic,liang2026evaluating}. Yet their ability to provide coherent responses across long interaction horizons remains fundamentally constrained by finite context windows and the difficulty of reliably using information across long contexts~\citep{bai2024longbench,liu2024lost}. Recent work augments the conversational agentic LLM systems with explicit \emph{external long-term memory}: a persistent store that encodes information from prior interactions and makes it available for later retrieval~\citep{zhang2024survey,wu2025human}.
Within this line of work, \emph{retrieval-augmented generation (RAG)}-based memory stores detailed records of past interactions as episodic memory traces in an external memory store, retrieves only a relevant subset at inference time using the user's utterance as a query, and conditions response generation on the retrieved traces. These approaches support responses that remain coherent with past interactions while keeping the input within a feasible context window. 

% Existing works on RAG-based memory 

% The central challenges of RAG-based memory are twofold: (1) how to
% encode, organize, and retrieve episodic traces from prior interactions,
% and (2) which queries to construct at retrieval time. Substantial
% progress has been made on the former by imposing increasingly structured
% organization in memory stores, including graph-structured memory
% organization and update methods such as Mem0~\citep{
% chhikara2025mem0}, MAGMA~\citep{jiang2026magma}, and A-MEM~\citep{
% xu2025amem}, as well as hierarchical tree-structured indexing methods such as
% MemTree~\citep{rezazadeh2024memtree} and RAPTOR~\citep{sarthi2024raptor}.
% The complementary question, however, remains underexplored: \emph{which queries should the agent form to retrieve the right memory trace?} 

%Substantial progress has been made 
Several works have focused on more effective encoding, organizing, and retrieving episodic traces, such as Mem0~\citep{
chhikara2025mem0}, MAGMA~\citep{jiang2026magma}, A-MEM~\citep{
xu2025amem}, and MemTree~\citep{rezazadeh2024memtree}.
However, %the complementary 
a key question remains underexplored: \emph{which queries should the agent form to retrieve the right memory trace?}. Existing methods merely focus on the surface form of the
utterance, relying on semantic similarity through global dense retrieval over
stored memory traces~\citep{sarthi2024raptor} or traversal heuristics over the predefined structures~\citep{xu2025amem,jiang2026magma}. However, to satisfy the goal of generating appropriate responses, conversational agentic systems require substantial reasoning over the memory~\citep{acikgoz-etal-2025-single}, and the information needed to support such reasoning is not always semantically adjacent to the utterance. Thus, existing approaches often fail on multi-hop questions that require chaining intermediate queries across memory entries.

% The central challenges of RAG-based memory are twofold: (1) how to
% encode, organize, and retrieve episodic traces from prior interactions,
% and (2) which queries to construct at retrieval time. Substantial
% progress has been made on the former by imposing increasingly structured
% organization in memory stores, including graph-structured memory
% organization and update methods such as Mem0~\citep{
% chhikara2025mem0}, MAGMA~\citep{jiang2026magma}, and A-MEM~\citep{
% xu2025amem}, as well as hierarchical tree-structured indexing methods such as
% MemTree~\citep{rezazadeh2024memtree} and RAPTOR~\citep{sarthi2024raptor}.
% The complementary question, however, remains underexplored: \emph{which queries should the agent form to retrieve the right memory trace?} Existing methods remain anchored to the surface form of the
% utterance, relying on semantic similarity through global dense retrieval over
% stored memory traces~\citep{sarthi2024raptor} or traversal heuristics over a
% predefined memory structure~\citep{xu2025amem,jiang2026magma}. Yet responses from conversational agentic systems often demand substantial reasoning~\citep{acikgoz-etal-2025-single}, and the information needed to support such reasoning is not always semantically adjacent to the current utterance. As a result, existing approaches often fail on multi-hop and commonsense questions that require chaining intermediate facts and latent relations across memory entries.

\begin{figure}
    \centering
    \includegraphics[width=\linewidth]{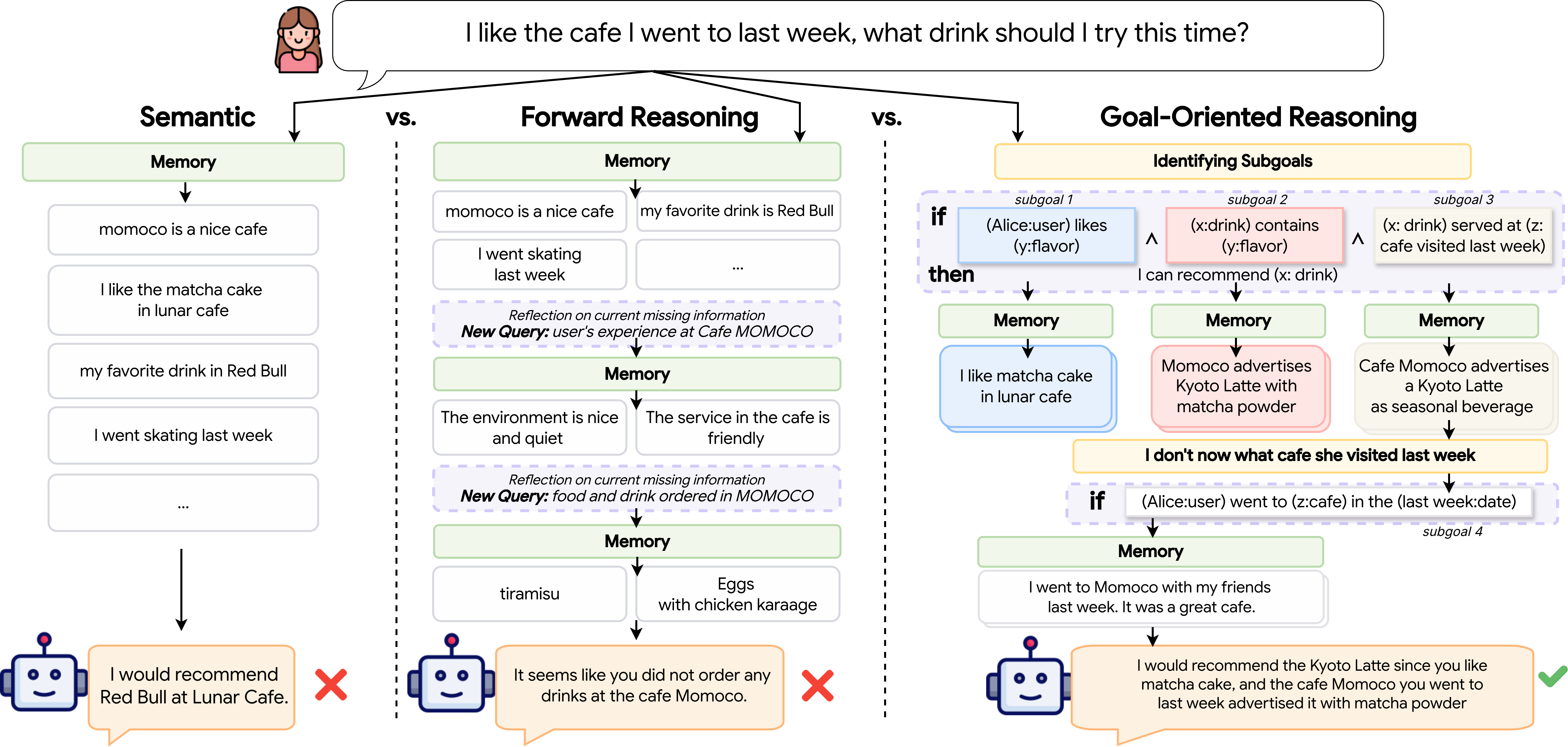}
\caption{Comparison of retrieval from external memory: utterance-based semantic retrieval (\emph{left}), forward reasoning (\emph{middle}), and \textsc{Goal-Mem} (\emph{right}). Utterance-based semantic retrieval and forward reasoning are easily misled by irrelevant memory traces. \textsc{Goal-Mem} decomposes the utterance into subgoals and performs targeted retrieval for memory-grounded verifiable response generation.}
    \label{fig:fig1_hook}
\end{figure}

Reasoning over external long-term memory poses several challenges. The agent must correctly identify intermediate queries to recover the required memory traces across multi-hop steps~\citep{trivedi2023interleaving,liu2025ma} and ground the final response in retrieved memory facts rather than relying on the model's parametric knowledge~\citep{niu2024ragtruth}. These challenges make many popular forward reasoning strategies, such as self-reflection and query reformulation, brittle: early irrelevant retrievals or unsupported inferences can propagate across steps, causing the agent to accumulate facts without converging on the correct set of memory traces, as illustrated in Figure~\ref{fig:fig1_hook} (middle). Such approaches also lack an explicit verification mechanism to ensure that the final answer is grounded in retrieved memory facts.

To address this, we propose \textsc{Goal-Mem}: \textbf{Goal}-Oriented
Reasoning for RAG-based \textbf{Mem}ory, a framework rooted
in the classical principle of backward chaining~\citep{fikes1971strips}, which can be applied to arbitrary memory backbones. Instead of expanding outward from
the utterance and the memory retrieved so far, \textsc{Goal-Mem}
treats the %utterance 
user's question as the goal and decomposes a set of intermediate subgoals whose satisfaction is necessary to establish the goal (cf. right of Figure~\ref{fig:fig1_hook}), and retrieves memory traces targeted at each subgoal. If the
retrieved memory is insufficient to resolve any subgoals, the agent then recursively derives their antecedents as queries, until the original
goal can be answered. To make this process verifiable, we represent both goals and
memory traces in typed \emph{Natural Language Logic} (NL-Logic). NL-Logic retains
a language-like expressiveness since predicates and types remain NL phrases, while inheriting the structural
guarantees of First-Order Logic (FOL) through typed variables and explicit
substitutions. As a key component of NL-Logic, we introduce a \emph{verifiable
unification} for memory where the subgoal is grounded only when the retrieved memory facts are type-consistent and logically entailed via LLM-judged entailment over NL predicates, ruling out the spurious pure semantic matches. 
In summary, our contributions are as follows:
\begin{itemize}
\item We propose \textbf{\textsc{Goal-Mem}}, a novel goal-oriented reasoning framework for RAG-based memory with NL-Logic. \textsc{Goal-Mem} performs backward chaining from the user utterance and retrieves only the goal-relevant facts with a verifiable response generation.

\item We show that \textsc{Goal-Mem} is structure-agnostic and can be seamlessly applied to popular RAG-based memory backbones, yielding substantial empirical improvements.

\item We compare \textsc{Goal-Mem} against popular forward reasoning approaches using the same memory backbones and show that \textsc{Goal-Mem} consistently yields larger gains.
\end{itemize}

\section{Related Work}
\label{related_work}
% \subsection{Retrieval-Augmented Generation (RAG) Based Memory}
\textbf{Retrieval-Augmented Generation (RAG) Based Memory} RAG-based memory retrieves a relevant subset of stored conversational traces or structured records at inference time. This paradigm has been widely studied in recent LLM-agent memory systems, where prior interactions are stored externally and selectively recalled to support long-term or task-grounded responses~\citep{zhang2024survey}. We therefore focus on representative retrieval-augmented memory backbones that cover the major design choices in this line of work. Early work like MemoryBank introduced updatable long-term memory with forgetting mechanisms \citep{zhong2023memorybank}. More recent systems focus on advanced structural organization and routing heuristics. For relational structures, Mem0 \citep{chhikara2025mem0} and A-MEM \citep{xu2025amem} utilize dynamic graphs and linked notes, concepts further advanced by agentic graph routing in AdaMem \citep{yan2026adamem} and causal tool retrieval in AMA-Agent \citep{zhao2026amabench}. To manage abstraction, memory is increasingly structured hierarchically; MemTree \citep{rezazadeh2024memtree} and RAPTOR \citep{sarthi2024raptor} construct multi-level trees. Furthermore, methods like MAGMA perform query-adaptive traversal across multi-view graphs \citep{jiang2026magma}.
However, \textsc{Goal-Mem} is orthogonal to all these works. Whereas prior methods primarily focus on memory structure and retrieval, the problem is addressing what information should be queried for the utterance. \textsc{Goal-Mem} serves as a structure-agnostic reasoning layer that can be flexibly integrated with these memory backbones to enforce goal-oriented verifiable reasoning.  

% \subsection{LLM Reasoning via Backward Chaining}
\textbf{LLM Reasoning via Backward Chaining} Backward chaining is an established paradigm in logical reasoning. %and theorem proving.
Unlike forward reasoning approaches such as query reformulation \citep{ma2023query}, self-reflection \citep{yao2022react, shinn2023reflexion}, iterative retrieval with LLM reranking~\citep{kim2026bayesian,liu2025multimodal}, and memory-specific methods such as MemGuide \citep{du2025memguide} that aim to derive the goal from observed facts progressively, backward chaining works backward from the goal, avoiding exhaustive exploration of facts and thereby yielding substantial efficiency gains~\citep{russell1995modern, al2015comparison}. This goal-directed approach has recently been used in LLM-based reasoning. LAMBADA~\cite{kazemi2023lambada} pioneered this integration by prompting the LLM to emulate the backward-chaining process by decomposing complex goals into %manageable 
sub-goals. Symba~\cite{lee2025symba} leverages a symbolic solver %reasoner 
that undergoes backward chaining and calls the LLM to provide %supplementary 
additional facts. %for the solver. 
LLM-TRes~\cite{toroghi2024verifiable} employs theory resolution to unify LLM-derived commonsense axioms into backward chaining. %Furthermore, Reason from Future~\cite{xu2025reason} imposes goal-oriented constraints on intermediate steps to reduce the search space in forward chain of thought reasoning. 

Despite the collective success of these works in enhancing LLM reasoning efficiency, they remain constrained by a reliance on static knowledge, and to our knowledge, no research has extended the potential of backward LLM reasoning to the dynamic domain of memory-augmented agents, where backward chaining must be tightly coupled with adaptive memory and continual context updates.
\section{Preliminaries}
\label{sec:preliminary}
\paragraph{Retrieval-Augmented Generation (RAG) Based Memory}
We consider a conversation as a sequence of turns. At turn $t$, the user provides an utterance $u_t$, and the agent generates a response $a_t$. Let $[(u_1, a_1), \ldots, (u_{t-1}, a_{t-1})]$ denote the \textit{dialogue history} prior to turn $t$. A RAG-based memory-augmented LLM is additionally equipped with an external long-term memory $\mathcal{M}$ that stores episodic traces distilled from dialogue history.

A RAG-based memory backbone $\mathcal{B}$ consists of three components: (1) a persistent external long-term memory store $\mathcal{M}$, instantiated as a flat vector database, graph-structured memory, or tree-structured memory; (2) a retrieval function $\mathcal{B}_{\text{retrieve}}(q_t, \mathcal{M})$ that returns memory facts from $\mathcal{M}$ given a query $q_t$; and (3) an update mechanism to encode information in $u_t$ into $\mathcal{M}$. 
Given the utterance $u_t$, the agent first constructs one or more queries $q_t$ that encapsulate the information needs in $u_t$. Each query is passed to the memory backbone $\mathcal{B}$, which retrieves relevant memory facts $\mathcal{M}_t$:
$
\mathcal{M}_t = \mathcal{B}_{\text{retrieve}}(q_t, \mathcal{M}).
$
The agent then generates a response conditioned on both the current utterance and the retrieved fact, $a_t = f^{\mathrm{LLM}}(u_t, \mathcal{M}_t)$. The backbone updates $\mathcal{M}$ with the new turn.

As noted in \autoref{related_work}, %decisions regarding 
choices on what information in dialogue is encoded, how it is organized and updated, and how queries are matched to traces in $\mathcal{M}$ are %governed 
made by $\mathcal{B}$ and fall outside the scope of our work. We focus on constructing a sequence of queries $q_t$ via goal-oriented reasoning. %backward chaining.
%Our focus is on how to construct a sequence of queries $q_t$ via goal-oriented backward chaining.

\paragraph{Natural Language Logic (NL-Logic)}
While conversation and memories are inherently expressed in NL, using LLMs for reasoning over pure NL induces the risk of generating ungrounded or hallucinated content. Moreover, NL-based reasoning is often difficult to verify, as the intermediate steps leading to a conclusion are not explicitly structured. Formal logical systems such as first-order logic (FOL)~\cite{russell1995modern} provide a structured framework with explicit variables, substitutions, and entailment relations, enabling transparent and verifiable reasoning. However, its reliance on predefined symbols limits its expressiveness for memory traces, motivating the need for a hybrid logical NL representation.
%Conversational memories are typically stored and retrieved as NL, which is expressive and naturally matches how conversational memories are stored and retrieved, but it often lacks type constraints and verifiable variable substitutions, which increases the risk of hallucinated response that not been grounded by retrieved memory. On the other fold, FOL provides explicit variables, substitutions, and entailment relations for stepwise verifiable reasoning, but its reliance on predefined predicates and type schemas makes it lack of the expressiveness for NL driven conversational memory.

To combine the strengths of both paradigms, we extend FOL into NL-Logic, where predicates, objects, and relations are expressed directly in NL, preserving logical structure while maintaining expressiveness.
%by expressing variables, objects, predicates, and functions in NL rather than as purely symbolic atoms with explicitly typed NL statements. 
Let $\mathcal{E}$ denote the constants appearing in the current utterance or retrieved memory facts, $\mathcal{V}$ a set of variables, and $\mathcal{T}$ the set of NL type descriptions.
%, and $\mathcal{P}_{\mathrm{NL}}$ the set of NL predicates. 
A typed term is written as $t:\tau$, where $t \in \mathcal{E}\cup\mathcal{V}$ and $\tau \in \mathcal{T}$. Unlike conventional typed logical systems, we do not assume that $\mathcal{T}$ is a fixed, predefined type vocabulary. Instead, types are open-ended, on-demand NL phrases and type checking is performed by the LLM through NL entailment. This allows more  flexible definitions.

In NL-Logic, an atomic formula is 
%retains the predicate--argument form,
% $
% p(t_1,\dots,t_n), \qxuad p \in \mathcal{P}_{\mathrm{NL}},
% $
% ,where $p$ denotes a NL predicate and each $t_i$ is a typed term.
% To preserve the expressivity and interpretability of natural language, we extend this representation as \textit{Extended NL-Logic} (NL-Logic):
$
\mathcal{R}(t_1,\dots,t_n),
$
where $\mathcal{R}$ is an NL relational phrase and each argument $t_i$ is explicitly typed (e.g., $\text{``}(Alice:\textit{user}) \textit{ likes } (x:\textit{flavor})\text{.''} $).
% In each NL-Logic  statement that corresponds to a relation:
% $\mathcal{R}(t_1, \dots, t_k) $,
Type constraints are enforced on variables:
$
x:\tau \Rightarrow x \in \mathcal{E}_\tau
$
meaning that $x$ belongs to $\mathcal{E}_\tau$, the subset of constants in $\mathcal{E}$ with type $\tau$. Type specification for constants (e.g., \textit{``Alice''}, \textit{``Matcha''}) is optional, as their names implicitly encode their semantic type. Thus, an atomic formula in NL-Logic is represented as a structured NL statement with typed arguments, rather than as a purely symbolic predicate application.
In this logical system, a knowledge base (e.g. Memory) can be represented as a set of ground atoms:
% $$ K = \{ \mathcal{R}(e_i, e_j)\; |(e_i,e_j) \subset \mathcal{E}^2\} $$
\begin{equation}
 \mathcal{M} = \{\mathcal{R}(e_1, \ldots, e_n) \mid (e_1, \ldots, e_n) \in \mathcal{E}^n\}
\end{equation}

% \begin{flowexample}
% \textbf{Example}:
% $$ K = \{ \text{``(\textit{Alice:user}) likes (\textit{Matcha: flavor})''} \} $$
% \end{flowexample}

% \textsc{Goal-Mem} focuses on determining the retrieval query $ Q_t$, namely what information should be queried from $\mathcal{M}$ via $\mathcal{B}$ to answer the current utterance $u_t$.
\section{Methodology}

Given a user utterance $u_t$ at turn $t$, the objective is to generate a grounded response. \textsc{Goal-Mem} first maps $u_t$ to a formal NL-Logic goal $\phi_{u_t}(x_1, \dots, x_m)$, which captures both the semantic intent of the utterance and the set of variables $\{x_1, \dots, x_m\}$ that must be instantiated to satisfy the user's request.

\begin{flowexample}{}
\textbf{Running Example 1.}\\
User utterance: $u_t = $ ``\textit{Alice: Which drink did I order at cafe Momoco?}''\\
NL-Logic goal: $\phi_{u_t} = $ ``(\textit{x:drink}) ordered by (\textit{Alice:human}) at (\textit{Momoco:cafe}).''
\end{flowexample}

Answering the user requires resolving variable $x$, whose type is \textit{drink}. The response-generation problem thus reduces to a logical inference: finding a substitution $\theta = \{x_1 / e_1, \dots, x_m / e_m\}$, where each variable $x_i$ is replaced by a type-consistent constant $e_i$ drawn from the retrieved memory $\mathcal{M}_t$.

\begin{flowexample}
% \textbf{Running Example 1 (continued).}\\
% $\mathcal{M}_t = \{$``(\textit{Alice}) ordered (\textit{Kyoto Latte}) at cafe Momoco.''$\}$\\
% $\theta = \{(\textit{x:drink}) / (\textit{Kyoto Latte})\}$
\textbf{Running Example 1 (continued).}\\
$\mathcal{M}_t = \{$``(\textit{Alice}) ordered (\textit{Kyoto Latte}) at cafe Momoco.''$\}$, \hspace{2pt}  
$\theta = \{(\textit{x:drink}) / (\textit{Kyoto Latte})\}$
\end{flowexample}

This formulation grounds the response in facts in retrieved memory, mitigating hallucinations that arise when existing approaches lack explicit verification on what fact should support the answer.

\subsection{Goal-Oriented Backward Chaining}

In practice, a direct substitution that satisfies $\phi_{u_t}$ using $\mathcal{M}_t$ rarely exists when $\phi_{u_t}$ requires multi-hop reasoning or open-domain knowledge. In these cases, \textsc{Goal-Mem} applies backward chaining to decompose $\phi_{u_t}$ into a series of subgoals, each of which can be independently grounded against $\mathcal{M}_t$.

\begin{figure}
    \centering
    \includegraphics[width=0.85\linewidth]{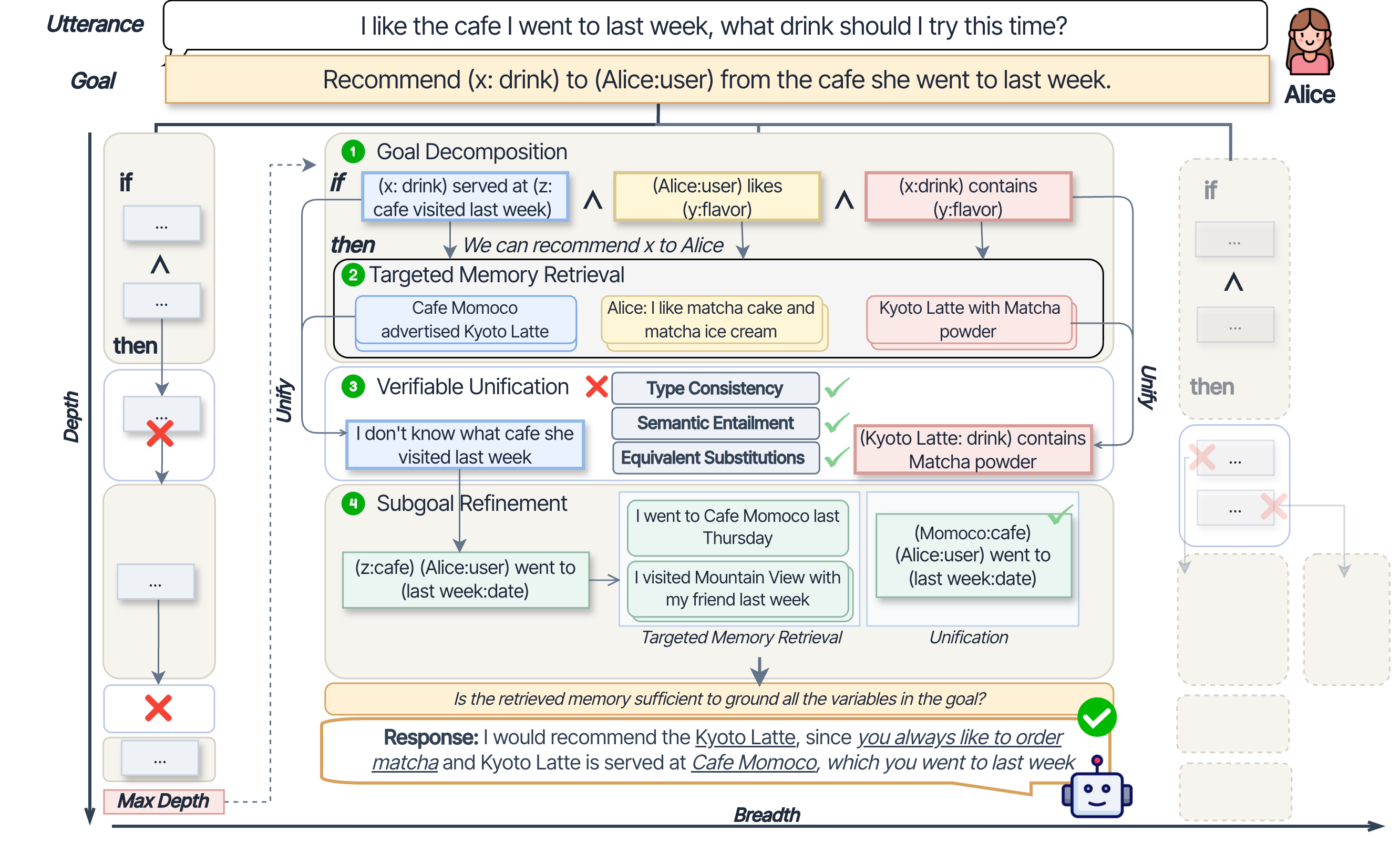}
    \caption{Overview of \textsc{Goal-Mem}. The framework starts from the user utterance and goal initialization (\emph{top}), decomposes the goal into NL-Logic subgoals for memory retrieval from a selected backbone (\emph{middle}), and checks whether the retrieved memory grounds all subgoals through unification. If not, it enters the depth loop (\emph{middle}), identifying new subgoals with targeted retrieval until all variables have been substituted or the maximum depth (\emph{bottom left}) is reached. The failed goal decomposition would then be provided to future decompositions to prevent a duplicate approach.}
    \label{fig:flowchart}
\end{figure}

As illustrated in Figure~\ref{fig:flowchart}, \textsc{Goal-Mem} operates through two nested iterative procedures. In each \emph{breadth loop}, \textsc{Goal-Mem} decomposes the user utterance $u_t$ into a set of atomic subgoals $\psi^{b}$ and retrieves the corresponding memory from the backbone $\mathcal{B}$ for each subgoal separately, yielding the aggregated retrieved memory $\mathcal{M}_t$. The goal is satisfied when all $\psi^{b}$ are grounded by $\mathcal{M}_t$ under a type-consistent non-conflicting substitution and all variables required by the goal are bound. If not, \textsc{Goal-Mem} refines each unresolved subgoal by inferring an intermediate subgoal whose satisfaction is necessary to establish the unresolved subgoal, and retrieves memory facts from $\mathcal{B}$. This \emph{depth loop} repeats until either every variable is successfully grounded or a maximum recursion depth is reached.
We initialize the backward chaining with an empty substitution $\theta^{(0)} = \emptyset$ and subgoal set $\psi^{(0)} = \emptyset$.

\paragraph{Goal Decomposition.}
At each breadth step $b \in \{1,\ldots, B_{\mathrm{max}}\}$, the agent maintains the current set of subgoals $\psi^{(b)}$, the retrieved memory $\mathcal{M}_t^{(b)}$, and the current substitution $\theta^{(b)}$. Given the original goal $\phi_{u_t}$, \textsc{Goal-Mem} decomposes the goal into a set of candidate subgoals:
\begin{equation}
\psi^{(b)} = \Psi_{\mathrm{decomp}}^{\mathrm{LLM}}\!\left(\phi_{u_t}, \{\psi^{(0)},\ldots,\psi^{(b-1)}\}\right),
\end{equation}
where $\Psi_{\mathrm{decomp}}^{\mathrm{LLM}}$ is an LLM-based decomposition function that avoids repeating earlier decompositions (i.e., $\{\psi^{(0)},\ldots,\psi^{(b-1)}\}$) that led to failed subgoals and instead proposes a distinct way to satisfy the goal. $\Psi_{\mathrm{decomp}}^{\mathrm{LLM}}$ induces a rule in NL-Logic of the form:
$
\psi_1 \land \cdots \land \psi_k \Rightarrow \phi_{u_t}, 
$
where each $\psi_i \in \psi^{(b)}$ is an atomic NL-Logic formula that must hold for $\phi_{u_t}$ to be satisfied.

\begin{flowexample}
\textbf{Running Example 2.}

$u_t = \text{``Alice: I like the cafe I went to last week, what drink should I try this time?''}$

$\phi_{u_t}=\text{``Recommend (\textit{x:drink}) to (\textit{Alice}) from the menu of the cafe she went to last week.''}$,

The LLM may generate the following rule:

\begin{adjustbox}{max width=\linewidth}
$\displaystyle
\text{``(\textit{Alice}) likes (\textit{y:flavor})''}
\land
\text{``(\textit{x:drink}) contains (\textit{y:flavor})''}
\land 
\text{``(\textit{x:drink}) served in (\textit{z:cafe visited last week})''}$
\end{adjustbox}

$\Rightarrow \text{``recommend (\textit{x:drink}) to (\textit{Alice}).''}$
\end{flowexample}

\paragraph{Targeted Memory Retrieval.}
Each subgoal $\psi_i \in \psi^{(b)}$, an atomic NL-Logic formula, is issued as a query to the memory backbone $\mathcal{B}$ to retrieve the corresponding facts:
\begin{equation}
\mathcal{M}_t^{(b)} = \bigcup_{\psi_i \in \psi^{(b)}} \mathcal{B}_{\text{retrieve}}(\psi_i,\mathcal{M}).
\end{equation}

\begin{flowexample}
\textbf{Running Example 2 (continued).}

For the subgoal $\psi_i = \text{``(\textit{x:drink}) contains (\textit{y:flavor})''},$ the memory backbone retrieves
\[
\mathcal{B}_{\text{retrieve}}(\psi_i,\mathcal{M}) = \left\{\text{``(\textit{Kyoto Latte}) with (\textit{Matcha Powder})''}\right\}.
\]
\end{flowexample}

\subsection{Verifiable Unification}

Given the subgoals $\psi^{(b)}$ and the aggregated retrieved memory $\mathcal{M}_t^{(b)}$, the agent grounds each subgoal by assessing whether the retrieved memories provide sufficient facts to satisfy it. Specifically, for each subgoal $\psi_i \in \psi^{(b)}$, the agent attempts to unify it with a retrieved memory fact $m_j \in \mathcal{M}_t^{(b)}$ conditioned on the current substitution $\theta^{(b)}$:
\begin{equation}
\delta_i = \operatorname{UNIFY}\!(\psi_i, m_j \mid \theta^{(b)}) \quad \text{where} \quad \operatorname{SUBST}(\theta^{(b)} \cup \delta_i,\psi_i) =_{\mathrm{LLM}} \operatorname{SUBST}(\theta^{(b)} \cup \delta_i, m_j).
\end{equation}

Here, $\delta_i$ denotes the candidate substitution induced by $m_j$ for $\psi_i$, and $\operatorname{SUBST}(\delta_i, p)$ denotes the formula obtained by applying substitution $\delta_i$ to $p$.

\begin{flowexample}
\textbf{Running Example 2 (continued).}
Recall a previous subgoal $\psi_i$ and a retrieved memory $m_j \in \mathcal{M}_t^{(b)}$:

\noindent\makebox[\linewidth][c]{%
\parbox{0.98\linewidth}{\centering
$\psi_i =$ ``(\textit{x:drink}) contains (\textit{y:flavor})''.}%
}\par
\noindent\makebox[\linewidth][c]{%
\parbox{0.98\linewidth}{\centering
$m_j =$ ``The cafe advertised a (\textit{Kyoto Latte}) with (\textit{Matcha Powder:ingredient})''.}%
}\par

To determine whether $\psi_i$ can be grounded by $m_j$, we compute $\mathrm{UNIFY}\!(\psi_i, m_j \mid \theta^{(b)})$.

$\mathrm{UNIFY}\!(\psi_i, m_j \mid \theta^{(b)})$ proposes a variable substitution:
$\delta_i = \{x \,/\, \text{``\textit{Kyoto Latte}''}\ , y \,/\, \text{``\textit{Matcha Powder}''}\}.$%

\end{flowexample}

Unification succeeds only if three conditions hold: (i) the proposed substitution satisfies type consistency; (ii) the proposed substitution is equivalent to the existing substitution $\theta^{(b)}$; and (iii) the retrieved memory fact $m_j$ logically entails the grounded subgoal. Because NL-Logic types and relations are expressed in natural language, these checks are performed by an LLM.

\paragraph{Type Consistency.}
A variable $(x:\tau)$ in the subgoal $\psi_i$ can be substituted with a constant $e'$ from a retrieved memory fact $m_j$ if and only if their types are consistent. If $e'$ is explicitly typed as $(e':\tau')$, we require $\tau' \vdash_{\mathrm{LLM}} \tau$; otherwise, the entity must be an instance of the type, i.e., $\tau_{\text{LLM}}(e' )$.

\begin{flowexample}
\textbf{Running Example 2 (continued).}

Since $\text{``\textit{drink}''}_{\text{LLM}}(\text{``\textit{Kyoto Latte}''}) $ and ``\textit{ingredient}'' $\vdash_{\mathrm{LLM}}$ ``\textit{flavor}'', 
the substitution $\delta_i$ is type-compatible.
% :

% \noindent\makebox[\linewidth][c]{%
% $\{x \,/\, \text{``\textit{Kyoto Latte}''}\}.$%
% }\par
\end{flowexample}

% \paragraph{Equivalence with Existing Substitutions.}
% The proposed substitution must not conflict with the previously identified substitution $\theta^{(b)}$. If a variable has already been grounded, a new substitution for the same variable is accepted only when it is semantically equivalent to the existing substitution under the current context. We denote this equality relation by $=_{\mathrm{LLM}}$, i.e.,

% \begin{flowexample}
% \textbf{Running Example.}
% Given $\theta^{(b)} = \{y \,/\, \text{``\textit{Matcha}''}\}$, $\mathrm{UNIFY}\!(\psi_i, m_j \mid \theta^{(b)})$ proposes a substitution:

% \noindent\makebox[\linewidth][c]{%
% $\delta_i^{+} = \{y \,/\, \text{``\textit{Matcha Powder}''}\}.$%
% }\par

% Since both $\delta_i^{+}$ and $\theta^{(b)}$ assign a value to $y$, we check for equality of the assigned substitutions and obtain: $\text{``\textit{Matcha}''} =_{\mathrm{LLM}} \text{``\textit{Matcha Powder}''}$.

% In contrast, if the proposed substitution is $\delta_i^{-} = \{y \,/\, \text{``\textit{Strawberry}''}\}$, then $\text{``\textit{Matcha}''} \neq_{\mathrm{LLM}}\text{``\textit{Strawberry}''}$ and the substitution is rejected.
% \end{flowexample}

\paragraph{Equality Checks with Existing Substitutions.}
 
If the substitution assigns a variable already bound in $\theta^{(b)}$, the assigned constants must be equal in the given context. Formally, let $\operatorname{dom}(\theta)$ denote the set of variables bound by a substitution $\theta$. For every variable $x \in \operatorname{dom}(\delta_i) \cap \operatorname{dom}(\theta^{(b)})$, the two bindings must agree under the LLM-judged equality relation $=_{\mathrm{LLM}}$:
\begin{equation}
\delta_i(x) =_{\mathrm{LLM}} \theta^{(b)}(x), \quad \forall x \in \operatorname{dom}(\delta_i) \cap \operatorname{dom}(\theta^{(b)}).
\end{equation}
% Variables bound by only one of the two substitutions impose no constraint.

\begin{flowexample}
\textbf{Running Example 2 (continued).}

Let $\theta^{(b)} = \{y \,/\, \text{``\textit{Matcha}''}\}$ be an existing substitution. Since
$y \in \operatorname{dom}(\delta_i) \cap \operatorname{dom}(\theta^{(b)})$ and $\text{``\textit{Matcha Powder}''} =_{\mathrm{LLM}} \text{``\textit{Matcha}''}$ in the given context, $\delta_i^{+}$ is accepted under this condition.

In contrast, if $\delta_i^{-} = \{y \,/\, \text{``\textit{Strawberry}''}\}$, then $\text{``\textit{Strawberry}''} \neq_{\mathrm{LLM}} \text{``\textit{Matcha}''}$, the substitution is rejected.
\end{flowexample}

\paragraph{Logical Entailment.}
We apply the type-consistent and substitution-consistent $\theta^{(b)} \cup \delta_i$ to obtain the grounded subgoal $\psi_i^{\mathrm{g}} = \psi_i[\theta^{(b)} \cup \delta_i]$. The LLM then checks whether any retrieved memory fact $m_j \in \mathcal{M}_t^{(b)}$ entails the grounded subgoal: $m_j \vdash_{\mathrm{LLM}} \psi_i^{\mathrm{g}}$.

\begin{flowexample}
\textbf{Running Example 2 (continued).}

Given the type-consistent and non-contradictory variable substitution

\noindent\makebox[\linewidth][c]{%
\parbox{0.98\linewidth}{\centering
$\delta_i = \{x \,/\, \text{``\textit{Kyoto Latte}''}, y \,/\, \text{``\textit{Matcha Powder}''}\},$}%
}\par

the subgoal $\psi_i$ is grounded to
$\psi_i^{\mathrm{g}} =$ ``(\textit{Kyoto Latte}) contains (\textit{Matcha Powder}).''% 
 Thus, $m_j \vdash_{\mathrm{LLM}} \psi_i^{\mathrm{g}}$.
\end{flowexample}

If all three conditions hold, the agent accepts the substitution $\delta_i$ inferred from $\psi_i$ and $m_j \in \mathcal{M}_t$, and updates $\theta^{(b)} \leftarrow \theta^{(b)} \cup \delta_i$. In practice, unification should be performed simultaneously across $\psi^{(b)}$ given $\mathcal{M}_t$ within a single LLM call to prevent bias in iteration order.

\subsection{Subgoal Refinement via Backward Chaining}

If any subgoals in $\psi^{(b)}$ fail to unify with $\mathcal{M}_t$, \textsc{Goal-Mem} enters the subgoal-refinement (depth) loop. At each depth step $d \in \{1,\ldots, D_{\mathrm{max}}\}$, \textsc{Goal-Mem} refines the unresolved subgoals as:
\begin{equation}
\psi^{(b,d)} = \{\Psi_{\mathrm{refine}}^{\mathrm{LLM}}(\psi_i, \mathcal{M}_t^{(b,d-1)}) : \psi_i \in \mathcal{U}^{(b,d-1)}\}.
\end{equation}
where $\psi^{(b,d-1)}$ denotes the set of subgoals that failed to unify at the previous depth step. The operator $\Psi_{\mathrm{refine}}^{\mathrm{LLM}}$ is an LLM-based refinement function that generates a requisite antecedent whose grounding is required to satisfy the unresolved subgoal. Given the refined subgoals $\psi^{(b,d)}_i = \Psi_{\mathrm{refine}}^{\mathrm{LLM}}(\psi^{(b,d-1)}_i)$, the agent then queries $\mathcal{B}$ to retrieve additional facts as $\mathcal{M}_t^{(b,d)} = \mathcal{M}_t^{(b,d-1)} \cup \mathcal{B}_{\text{retrieve}}\!(\psi^{(b,d)}, \mathcal{M})$. The agent then re-applies the unification procedure over $\mathcal{M}_t^{(b,d)}$ to ground the unresolved subgoals.
This process continues until all original and newly identified subgoals can be grounded by the accumulated retrieved memory $\mathcal{M}_t^{(b,d)}$, no new memory can be retrieved, or maximum depth is reached. 

\begin{flowexample}
\small
\textbf{Running Example 2 (continued).}

\noindent\makebox[\linewidth][c]{%
$\psi_i^{(b,d-1)} = \text{``$(x:\textit{drink})$ served in $(z:\textit{cafe\ visited\ last\ week})$''},$%
}

\noindent\makebox[\linewidth][c]{%
$m_j = \text{``The $(\textit{Momoco cafe})$ advertised a $(\textit{Kyoto\ Latte})$ as its seasonal drink.''}$%
}

Unification fails ($\textit{``Momoco cafe''} \;\nvdash_{\mathrm{LLM}}\; \textit{``cafe\ visited\ last\ week''}$) since the cafe visited last week is unknown.

\noindent\makebox[\linewidth][c]{$\Psi_{\mathrm{refine}}^{\mathrm{LLM}}\!(\psi_i^{(b,d-1)}) = \psi_i^{(b,d)} = \text{``$(\textit{Alice})$ visited $(z:\textit{cafe})$ last week.''}$
}
\end{flowexample}

If no new memory can be retrieved or the maximum depth is reached, this decomposition is considered infeasible for the given $\mathcal{M}$ and $\mathcal{B}$.. The system then generates a different decomposition $\psi^{(b+1)}$ of the original goal. This process continues until the maximum number of breadth steps.

\subsection{Answer Generation and Memory Updates}

If verifiable unification grounds all active subgoals and binds every required answer variable in the original goal, yielding a grounded goal $\phi_{u_t}^{g}$ supported by retrieved memory facts, the goal is considered satisfied. The agent then generates the final response by conditioning an LLM on the user utterance $u_t$, the grounded goal $\phi_{u_t}^{g}$, and the supporting facts $\mathcal{F}$ used to ground the subgoals.
The new turn, or information distilled from it, is written to the persistent memory $\mathcal{M}$ directly via $\mathcal{B}$, as defined in Section~\ref{sec:preliminary}. \textsc{Goal-Mem} therefore inherits the storage and update semantics of any compatible backbone without modification. We provide the complete algorithm in \autoref{app:algorithm}. 
% Here are the final grounded goals for the example.
\begin{flowexample}
\textbf{Running Example 2 (continued).}

$f^{\mathrm{LLM}}(u_t, \phi_{u_t}^g, \mathcal{F})$= \textit{``I recommend the Kyoto Latte, a matcha-flavored drink from Cafe Momoco''}
\end{flowexample}

\subsection{Complexity}
\textsc{Goal-Mem} requires $\mathcal{O}(B_{\mathrm{max}}D_{\mathrm{max}})$ LLM calls and $\mathcal{O}(B_{\mathrm{max}}D_{\mathrm{max}}K_{\mathrm{max}})$ retrievals per turn. $K_{\mathrm{max}}$ be an upper bound on the number of subgoals at any depth step. In practice, the cost is significantly lower since execution often terminates early (i.e., within $B{=}1$, $D{\leq}3$ as we show in \autoref{appendix:complexity}), and batched unification keeps LLM calls per depth step constant. As a result, its token cost is comparable to forward-reasoning methods. (cf. Appendix~\ref{appendix:complexity} for additional details and token costs analysis).

% Let $K_{\mathrm{max}}$ be an upper bound on the number of subgoals at any depth step. We exclude memory backbone LLM calls, since the latter depend on the implementation of $\mathcal{B}_{\mathrm{retrieve}}$. For each breadth attempt, \textsc{Goal-Mem} uses one LLM call for goal decomposition and one batched LLM call for verifiable unification over all retrieved facts and active subgoals. At each refinement depth, it then uses one LLM call to refine the unresolved subgoals and one batched LLM call to re-run unification after the additional retrieval. Thus, the LLM calls is bounded by $1 + B_{\mathrm{max}}\bigl(2 + 2D_{\mathrm{max}}\bigr)$ or $\mathcal{O}(B_{\mathrm{max}}D_{\mathrm{max}})$ LLM calls. 
% The memory retrieval is bounded by:
% $
% C_{\mathrm{ret}} \leq B_{\mathrm{max}}(D_{\mathrm{max}}+1)K_{\mathrm{max}}.
% $

% We conduct various empirical analysis in \autoref{fig:ablation} and \autoref{app:realized-search-distribution}, in practice, the realized cost is often lower because it frequently terminates before the algorithm reaches $B_{\mathrm{max}}$ and $D_{\mathrm{max}}$ (i.e., 95 percentiles of samples have finished within $B = 1$ and $D = 3$ across datasets), as soon as all variables in $\phi_{u_t}$ are grounded. In addition, under batched unification, \textsc{Goal-Mem} uses only a constant number of LLM calls per depth step. Thus, since $B_{\mathrm{max}}$ and $D_{\mathrm{max}}$ are small, \textsc{Goal-Mem} has a similar asymptotic per-turn LLM-call complexity as existing forward-reasoning approaches.

\section{Experiments}
To evaluate the effectiveness of the proposed \textsc{Goal-Mem}
%\footnote{\url{https://anonymous.4open.science/r/goal-mem-6A8A/README.md}} 
backward chaining framework on real agentic memory tasks, we consider the following research questions:

\textbf{RQ1:} How does \textsc{Goal-Mem}, when paired with existing memory backbones, compare with each memory backbone alone?

\textbf{RQ2:} How does \textsc{Goal-Mem} compare with forward-reasoning approaches (e.g., Query Reformulation, Self-Reflection, MemGuide, and ReAct) under matched memory-backbone settings?

% \textbf{RQ3:} How effective is the proposed goal-directed backward chaining approach in leveraging the LLM's commonsense knowledge to connect facts to what's needed to answer the question?

% \textbf{RQ3:} How effective is the proposed goal-directed backward chaining approach in reducing LLM hallucinations?

\textbf{RQ3:} How do different parameters, such as the maximum breadth and depth. of \textsc{Goal-Mem} reasoning, contribute to overall performance?

\providecolor{gormblue}{RGB}{235,243,255}

\begin{table*}[t]
\centering
\small
\setlength{\tabcolsep}{2.6pt}
\renewcommand{\arraystretch}{1.15}
\resizebox{\linewidth}{!}{%
\begin{tabular}{lll cc >{\columncolor{gormblue}}c>{\columncolor{gormblue}}c>{\columncolor{gormblue}}c>{\columncolor{gormblue}}c cc >{\columncolor{gormblue}}c>{\columncolor{gormblue}}c>{\columncolor{gormblue}}c>{\columncolor{gormblue}}c}
\toprule
\multirow{3}{*}{\textbf{Dataset}}
& \multirow{3}{*}{\textbf{Category}}
& \multirow{3}{*}{\textbf{Backbone}}
& \multicolumn{6}{c}{\textbf{Gemma-4-26B}}
& \multicolumn{6}{c}{\textbf{GPT-5.4-mini}} \\
\cmidrule(lr){4-9} \cmidrule(lr){10-15}
& &
& \multicolumn{2}{c}{\textbf{Backbone Alone}}
& \multicolumn{4}{c}{\cellcolor{gormblue}\textbf{Backbone with \textsc{Goal-Mem}}}
& \multicolumn{2}{c}{\textbf{Backbone Alone}}
& \multicolumn{4}{c}{\cellcolor{gormblue}\textbf{Backbone with \textsc{Goal-Mem}}} \\
\cmidrule(lr){4-5} \cmidrule(lr){6-9} \cmidrule(lr){10-11} \cmidrule(lr){12-15}
& &
& \textbf{LLM Acc.\,(\%)} & \textbf{F1} & \textbf{LLM Acc.\,(\%)} & \textbf{$\Delta$Acc.\,(\%)} & \textbf{F1} & \textbf{$\Delta$F1} & \textbf{LLM Acc.\,(\%)} & \textbf{F1} & \textbf{LLM Acc.\,(\%)} & \textbf{$\Delta$Acc.\,(\%)} & \textbf{F1} & \textbf{$\Delta$F1} \\
\midrule

\multirow{9}{*}{\textbf{LoCoMo}} & \multirow{3}{*}{\textbf{No Structure}} & RAG (Dense) & 62.39\,{\scriptsize (2.42)} & 0.543\,{\scriptsize (0.018)} & 79.44\,{\scriptsize (2.02)} & $\uparrow$17.06 & 0.618\,{\scriptsize (0.018)} & $\uparrow$0.075 & 63.10\,{\scriptsize (2.41)} & 0.519\,{\scriptsize (0.019)} & 76.10\,{\scriptsize (2.13)} & $\uparrow$13.00 & 0.584\,{\scriptsize (0.019)} & $\uparrow$0.065 \\
 &  & RAG (BM25) & 60.51\,{\scriptsize (2.44)} & 0.530\,{\scriptsize (0.019)} & 77.69\,{\scriptsize (2.08)} & $\uparrow$17.19 & 0.610\,{\scriptsize (0.018)} & $\uparrow$0.080 & 66.86\,{\scriptsize (2.35)} & 0.542\,{\scriptsize (0.019)} & 74.58\,{\scriptsize (2.17)} & $\uparrow$7.72 & 0.571\,{\scriptsize (0.019)} & $\uparrow$0.029 \\
 &  & Mem0 & 61.93\,{\scriptsize (2.42)} & 0.484\,{\scriptsize (0.019)} & 68.22\,{\scriptsize (2.32)} & $\uparrow$6.29 & 0.508\,{\scriptsize (0.019)} & $\uparrow$0.024 & 63.42\,{\scriptsize (2.40)} & 0.505\,{\scriptsize (0.019)} & 65.95\,{\scriptsize (2.37)} & $\uparrow$2.53 & 0.501\,{\scriptsize (0.019)} & $\downarrow$0.004 \\
\cmidrule(lr){2-15}
 & \multirow{2}{*}{\textbf{Tree-Based}} & MemTree & 42.41\,{\scriptsize (2.47)} & 0.332\,{\scriptsize (0.018)} & 54.60\,{\scriptsize (2.49)} & $\uparrow$12.19 & 0.414\,{\scriptsize (0.020)} & $\uparrow$0.082 & 44.62\,{\scriptsize (2.48)} & 0.388\,{\scriptsize (0.019)} & 60.51\,{\scriptsize (2.44)} & $\uparrow$15.89 & 0.468\,{\scriptsize (0.020)} & $\uparrow$0.081 \\
 &  & RAPTOR & 51.10\,{\scriptsize (2.50)} & 0.441\,{\scriptsize (0.019)} & 65.30\,{\scriptsize (2.38)} & $\uparrow$14.20 & 0.519\,{\scriptsize (0.019)} & $\uparrow$0.078 & 55.90\,{\scriptsize (2.48)} & 0.478\,{\scriptsize (0.019)} & 63.23\,{\scriptsize (2.41)} & $\uparrow$7.33 & 0.521\,{\scriptsize (0.019)} & $\uparrow$0.042 \\
\cmidrule(lr){2-15}
 & \multirow{3}{*}{\textbf{Graph-Based}}  & Mem0-Graph & 59.53\,{\scriptsize (3.23)} & 0.450\,{\scriptsize (0.025)} & 63.36\,{\scriptsize (3.17)} & $\uparrow$3.83 & 0.468\,{\scriptsize (0.025)} & $\uparrow$0.018 & 61.93\,{\scriptsize (2.42)} & 0.469\,{\scriptsize (0.019)} & 62.91\,{\scriptsize (2.41)} & $\uparrow$0.97 & 0.463\,{\scriptsize (0.019)} & $\downarrow$0.006 \\
 &  & A-MEM & 44.42\,{\scriptsize (2.48)} & 0.386\,{\scriptsize (0.019)} & 59.08\,{\scriptsize (2.45)} & $\uparrow$14.66 & 0.457\,{\scriptsize (0.020)} & $\uparrow$0.071 & 41.76\,{\scriptsize (2.46)} & 0.378\,{\scriptsize (0.019)} & 65.05\,{\scriptsize (2.38)} & $\uparrow$23.28 & 0.523\,{\scriptsize (0.020)} & $\uparrow$0.145 \\
 &  & GraphRAG & 55.51\,{\scriptsize (2.48)} & 0.456\,{\scriptsize (0.019)} & 65.82\,{\scriptsize (2.37)} & $\uparrow$10.31 & 0.510\,{\scriptsize (0.019)} & $\uparrow$0.054 & 57.91\,{\scriptsize (2.46)} & 0.476\,{\scriptsize (0.019)} & 64.50\,{\scriptsize (2.28)} & $\uparrow$6.59 & 0.509\,{\scriptsize (0.018)} & $\uparrow$0.033 \\
\midrule
\multirow{9}{*}{\textbf{LongMemEval}} & \multirow{3}{*}{\textbf{No Structure}} & RAG (Dense) & 53.33\,{\scriptsize (5.56)} & 0.165\,{\scriptsize (0.019)} & 63.00\,{\scriptsize (5.46)} & $\uparrow$9.67 & 0.496\,{\scriptsize (0.049)} & $\uparrow$0.331 & 58.89\,{\scriptsize (4.19)} & 0.205\,{\scriptsize (0.030)} & 61.67\,{\scriptsize (5.10)} & $\uparrow$2.78 & 0.505\,{\scriptsize (0.064)} & $\uparrow$0.299 \\
 &  & RAG (BM25) & 55.33\,{\scriptsize (5.63)} & 0.160\,{\scriptsize (0.020)} & 64.00\,{\scriptsize (5.43)} & $\uparrow$8.67 & 0.503\,{\scriptsize (0.049)} & $\uparrow$0.343 & 54.44\,{\scriptsize (4.28)} & 0.210\,{\scriptsize (0.031)} & 57.78\,{\scriptsize (5.22)} & $\uparrow$3.33 & 0.491\,{\scriptsize (0.064)} & $\uparrow$0.280 \\
 &  & Mem0 & 49.67\,{\scriptsize (5.66)} & 0.200\,{\scriptsize (0.022)} & 58.67\,{\scriptsize (5.57)} & $\uparrow$9.00 & 0.457\,{\scriptsize (0.049)} & $\uparrow$0.257 & 54.17\,{\scriptsize (4.92)} & 0.202\,{\scriptsize (0.032)} & 59.17\,{\scriptsize (7.79)} & $\uparrow$5.00 & 0.484\,{\scriptsize (0.079)} & $\uparrow$0.282 \\
\cmidrule(lr){2-15}
 & \multirow{2}{*}{\textbf{Tree-Based}} & MemTree & 56.00\,{\scriptsize (5.62)} & 0.155\,{\scriptsize (0.018)} & 66.33\,{\scriptsize (5.35)} & $\uparrow$10.33 & 0.505\,{\scriptsize (0.048)} & $\uparrow$0.350 & 51.67\,{\scriptsize (5.27)} & 0.208\,{\scriptsize (0.026)} & 55.83\,{\scriptsize (5.28)} & $\uparrow$4.16& 0.436\,{\scriptsize (0.055)} & $\uparrow$0.228 \\
 &  & RAPTOR & 51.11\,{\scriptsize (5.30)} & 0.204\,{\scriptsize (0.031)} & 65.67\,{\scriptsize (5.37)} & $\uparrow$14.56 & 0.512\,{\scriptsize (0.049)} & $\uparrow$0.308 & 50.00\,{\scriptsize (4.16)} & 0.216\,{\scriptsize (0.031)} & 56.67\,{\scriptsize (5.24)} & $\uparrow$6.67 & 0.487\,{\scriptsize (0.064)} & $\uparrow$0.271 \\
\cmidrule(lr){2-15}
 & \multirow{3}{*}{\textbf{Graph-Based}}  & Mem0-Graph & 44.67\,{\scriptsize (5.63)} & 0.138\,{\scriptsize (0.015)} & 51.00\,{\scriptsize (5.66)} & $\uparrow$6.33 & 0.409\,{\scriptsize (0.049)} & $\uparrow$0.271 & 48.33\,{\scriptsize (4.94)} & 0.194\,{\scriptsize (0.039)} & 52.16\,{\scriptsize (5.94)} & $\uparrow$3.83 & 0.410\,{\scriptsize (0.078)} & $\uparrow$0.217 \\
 &  & A-MEM & 56.33\,{\scriptsize (5.51)} & 0.166\,{\scriptsize (0.019)} & 64.00\,{\scriptsize (5.43)} & $\uparrow$7.67 & 0.497\,{\scriptsize (0.049)} & $\uparrow$0.331 & 52.50\,{\scriptsize (5.84)} & 0.204\,{\scriptsize (0.036)} & 59.17\,{\scriptsize (5.79)} & $\uparrow$6.67 & 0.493\,{\scriptsize (0.080)} & $\uparrow$0.290 \\
 &  & GraphRAG & 53.00\,{\scriptsize (5.65)} & 0.160\,{\scriptsize (0.019)} & 61.67\,{\scriptsize (5.50)} & $\uparrow$8.67 & 0.472\,{\scriptsize (0.050)} & $\uparrow$0.312 & 46.33\,{\scriptsize (4.94)} & 0.198\,{\scriptsize (0.038)} & 50.83\,{\scriptsize (5.94)} & $\uparrow$4.50 & 0.431\,{\scriptsize (0.080)} & $\uparrow$0.234 \\
\bottomrule
\end{tabular}%
}
\caption{RQ1 cross-backbone comparison of standalone memory backbones and the same backbones augmented with \textsc{Goal-Mem}. $\Delta$Acc. and $\Delta$F1 report the absolute improvement over the corresponding backbone-alone result. Values in parentheses are 95\% confidence-interval half-widths .}
\label{tab:rq1_overall}
\end{table*}

% \begin{figure}
%     \centering
%     \includegraphics[width=\linewidth]{img/rq1_overall_goalmem_1x6.pdf}
%     \caption{RQ1 cross-backbone comparison of standalone memory backbones and the same backbones augmented with \textsc{Goal-Mem}. $\Delta$Acc. and $\Delta$F1 reports the absolute improvement over the corresponding backbone-alone result. \textsc{Goal-Mem} columns are highlighted in light blue. Values in parentheses are 95\% confidence-interval half-widths.}
%     \label{fig:rq1:overall}
% \end{figure}

% \subsection{Experimental Setup}

\paragraph{Memory Backbones and Baselines.}
We instantiate \textsc{Goal-Mem} with eight retrieval-augmented memory backbones: BM25-RAG~\citep{robertson2009probabilistic}, Dense-RAG~\citep{karpukhin2020dense}, Mem0 and Mem0-Graph~\citep{chhikara2025mem0}, MemTree~\citep{rezazadeh2024memtree}, RAPTOR~\citep{sarthi2024raptor}, GraphRAG~\citep{edge2024local}, and A-MEM~\citep{xu2025amem}. These cover unstructured, tree-based, and graph-based memory designs. Since public implementations differ in availability and interfaces, we implement all methods in a unified evaluation pipeline following their original algorithmic descriptions, while controlling for retrieval budgets, prompts, and backbone.
We compare against forward reasoning baselines: query reformulation~\citep{ma2023query}, LLM-based self-reflection, MemGuide~\citep{du2025memguide}, and ReAct~\citep{yao2022react} under the same memory backbone $\mathcal{B}$. Please refer to \autoref{app:forward-baselines} for details.

\paragraph{Datasets and Setup.}

We evaluate \textsc{Goal-Mem} on two long-term conversational memory benchmarks: LoCoMo~\citep{maharana2024evaluating} and LongMemEval-Small~\citep{wu2024longmemeval}. LoCoMo emphasizes long-range, multi-session memory with many multi-hop questions over dispersed evidence, while LongMemEval evaluates core assistant memory abilities, including extraction, multi-session and temporal reasoning, knowledge updates, and abstention (cf. \autoref{appendix:datasets} for details).
We use two LLM backbones: \texttt{GPT-5.4 mini}~\citep{openai2026gpt54mini} and \texttt{Gemma-4-26B-A4B}~\citep{google2026gemma4}.
We use identical hyperparameters in the standalone and \textsc{Goal-Mem}-augmented settings. The performance delta ($\Delta$) isolates the contribution of \textsc{Goal-Mem} from backbone-specific tuning. We maintain a consistent retrieval budget across methods: the memory backbone alone retrieves up to $k=60$ memories in a single call, while reasoning-based approaches distribute the same retrieval cap across iterations. The maximum breadth is consistently set to $3$, and the maximum depth is set to $5$. We evaluate the generated responses with reference answer via LLM-as-judge using prompts provided by LoCoMo and LongMemEval, and token F1. 
\subsection{Experimental Results}
\label{sec:results}

\paragraph{RQ1.}
\autoref{tab:rq1_overall} reports performance with and without \textsc{Goal-Mem} across two datasets and a range of memory backbones. \textsc{Goal-Mem} delivers consistent gains in both LLM-as-judge accuracy and token-level F1, regardless of how the underlying memory is organized. The largest absolute improvements appear on the simplest flat backbones, BM25-RAG and Dense-RAG, suggesting that goal-oriented backward chaining can compensate for limited memory structure. Mem0 and Mem0-Graph show smaller gains because their abstract memory facts omit fine-grained cues such as temporal details and raw dialogue context, weakening intermediate subgoal retrieval, which can be mitigated through a modification in structures, but altering the original backbone designs.

\begin{figure}[!h]
    \centering
    \includegraphics[width=\linewidth]{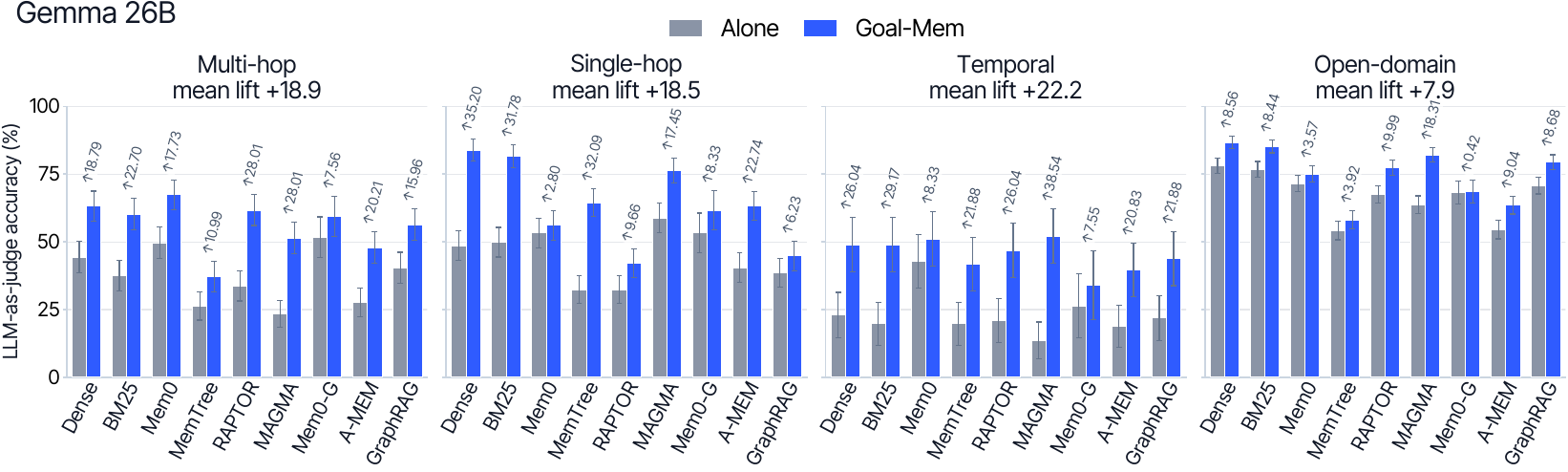}
    \caption{LLM accuracy by question type on LoCoMo with \texttt{Gemma-4-26B}. \textsc{Goal-Mem} consistently improves multi-hop performance across all backbones. (cf. \autoref{app:locomo_by_type} for \texttt{GPT-5.4 mini} result).}
    \label{fig:rq1:question_type}
\end{figure}

To localize the source of these gains, Figure~\ref{fig:rq1:question_type} breaks down accuracy by question type on LoCoMo. \textsc{Goal-Mem} shows a substantial lift on multi-hop questions, even for backbones whose overall improvements are smaller. This pattern supports the claim that semantic similarity alone is insufficient for multi-hop questions because the intermediate queries needed for chaining are often missing. \textsc{Goal-Mem}'s goal-directed backward chaining directly addresses this problem.

\paragraph{RQ2.}
Table~\ref{tab:rq2_reasoning_comparison} compares \textsc{Goal-Mem} with representative forward reasoning approaches under identical LLM and memory backbone settings. \textsc{Goal-Mem} either outperforms or achieves similar LLM accuracy the forward reasoning baseline across most backbones, including memory-tailored forward methods such as MemGuide. Our qualitative analysis (cf.\ \autoref{app:qualitative-multihop-failures}) further shows that \textsc{Goal-Mem} achieves the large gains on multi-hop questions, where forward baselines often retrieve relevant memory facts but fail to utilize them across hops to produce a valid response. \textsc{Goal-Mem} mitigates this by unifying and grounding subgoals with their corresponding memory facts prior to response generation, providing more guided chaining across hops for correct retrieval utilization.

\providecolor{gormblue}{RGB}{235,243,255}
\begin{table*}[t]
\centering
\small
\setlength{\tabcolsep}{3.0pt}
\renewcommand{\arraystretch}{1.15}
\resizebox{\linewidth}{!}{%
\begin{tabular}{lll c c c c >{\columncolor{gormblue}}c c c c c >{\columncolor{gormblue}}c}
\toprule
\multirow{2}{*}{\textbf{Dataset}}
& \multirow{2}{*}{\textbf{Category}}
& \multirow{2}{*}{\textbf{Backbone}}
& \multicolumn{5}{c}{\textbf{Gemma-4-26B}\,(\%)}
& \multicolumn{5}{c}{\textbf{GPT-5.4-mini}\,(\%)} \\
\cmidrule(lr){4-8} \cmidrule(lr){9-13}
& & & \textbf{Query Reformulation} & \textbf{Self-Reflection} & \textbf{MemGuide} & \textbf{ReAct} & \cellcolor{gormblue}\textbf{\textsc{Goal-Mem}} & \textbf{Query Reformulation} & \textbf{Self-Reflection} & \textbf{MemGuide} & \textbf{ReAct} & \cellcolor{gormblue}\textbf{\textsc{Goal-Mem}} \\
\midrule
\multirow{9}{*}{\textbf{LoCoMo}} & \multirow{3}{*}{\textbf{No Structure}} & RAG (Dense) & 56.49\,{\scriptsize (2.47)} & 56.23\,{\scriptsize (2.48)} & 57.72\,{\scriptsize (2.47)} & 44.81\,{\scriptsize (2.48)} & \textbf{79.44}\,{\scriptsize (2.02)} & 66.60\,{\scriptsize (2.35)} & 67.32\,{\scriptsize (2.34)} & 67.96\,{\scriptsize (2.33)} & 41.05\,{\scriptsize (2.60)} & \textbf{76.10}\,{\scriptsize (2.13)} \\
 &  & RAG (BM25) & 60.89\,{\scriptsize (2.44)} & 60.51\,{\scriptsize (2.44)} & 60.44\,{\scriptsize (2.44)} & 50.45\,{\scriptsize (2.50)} & \textbf{77.69}\,{\scriptsize (2.08)} & 71.85\,{\scriptsize (2.24)} & 70.69\,{\scriptsize (2.27)} & 69.58\,{\scriptsize (2.30)} & 45.46\,{\scriptsize (2.49)} & \textbf{74.58}\,{\scriptsize (2.17)} \\
 &  & Mem0 & 57.72\,{\scriptsize (2.47)} & 58.04\,{\scriptsize (2.46)} & 57.72\,{\scriptsize (2.47)} & 64.51\,{\scriptsize (2.91)} & \textbf{68.22}\,{\scriptsize (2.32)} & 62.52\,{\scriptsize (2.42)} & \textbf{66.02}\,{\scriptsize (2.36)} & 63.88\,{\scriptsize (2.40)} & 63.04\,{\scriptsize (2.41)} & 65.95\,{\scriptsize (2.37)} \\
\cmidrule(lr){2-13}
 & \multirow{2}{*}{\textbf{Tree-Based}} & MemTree & 47.73\,{\scriptsize (2.49)} & 46.43\,{\scriptsize (2.49)} & 44.81\,{\scriptsize (2.48)} & 51.59\,{\scriptsize (3.04)} & \textbf{54.60}\,{\scriptsize (2.49)} & 56.94\,{\scriptsize (2.47)} & 56.94\,{\scriptsize (2.47)} & 55.13\,{\scriptsize (1.56)} & 54.33\,{\scriptsize (2.11)} & \textbf{60.51}\,{\scriptsize (2.44)} \\
 &  & RAPTOR & 54.73\,{\scriptsize (2.48)} & 52.33\,{\scriptsize (2.49)} & 52.79\,{\scriptsize (2.49)} & 51.98\,{\scriptsize (3.04)} & \textbf{65.30}\,{\scriptsize (2.38)} & 54.35\,{\scriptsize (2.49)} & 58.24\,{\scriptsize (2.46)} & 59.14\,{\scriptsize (2.45)} & 52.97\,{\scriptsize (4.97)} & \textbf{63.23}\,{\scriptsize (2.41)} \\
\cmidrule(lr){2-13}
 &  \multirow{3}{*}{\textbf{Graph-Based}} & Mem0-Graph & 55.36\,{\scriptsize (3.27)} & 55.69\,{\scriptsize (3.27)} & 52.56\,{\scriptsize (3.04)} & 56.80\,{\scriptsize (3.01)} & \textbf{63.36}\,{\scriptsize (3.17)} & 58.55\,{\scriptsize (2.97)} & 57.76\,{\scriptsize (3.01)} & \textbf{63.68}\,{\scriptsize (4.70)} & 52.16\,{\scriptsize (1.12)} & 62.91\,{\scriptsize (2.41)} \\
 &  & A-MEM & 38.85\,{\scriptsize (2.43)} & 36.38\,{\scriptsize (2.40)} & 34.85\,{\scriptsize (2.66)} & 35.29\,{\scriptsize (2.91)} & \textbf{59.08}\,{\scriptsize (2.45)} & 48.64\,{\scriptsize (2.49)} & 56.42\,{\scriptsize (2.47)} & 52.59\,{\scriptsize (2.49)} & 46.16\,{\scriptsize (2.68)} & \textbf{65.05}\,{\scriptsize (2.38)} \\
 &  & GraphRAG & 55.97\,{\scriptsize (2.48)} & 54.99\,{\scriptsize (2.48)} & 56.03\,{\scriptsize (2.48)} & 37.99\,{\scriptsize (2.95)} & \textbf{65.82}\,{\scriptsize (2.37)} & 60.57\,{\scriptsize (2.44)} & 60.05\,{\scriptsize (2.44)} & 53.31\,{\scriptsize (2.49)} & 46.54\,{\scriptsize (2.16)} & \textbf{64.50}\,{\scriptsize (2.28)} \\
\midrule
\multirow{9}{*}{\textbf{LongMemEval}} & \multirow{3}{*}{\textbf{No Structure}} & RAG (Dense) & 58.33\,{\scriptsize (5.58)} & 57.67\,{\scriptsize (5.59)} & 59.67\,{\scriptsize (5.55)} & 45.00\,{\scriptsize (4.27)} & \textbf{63.00}\,{\scriptsize (5.46)} & 58.33\,{\scriptsize (4.20)} & 58.89\,{\scriptsize (4.19)} & \textbf{62.22}\,{\scriptsize (7.08)} & 35.56\,{\scriptsize (6.99)} & 61.67\,{\scriptsize (5.10)} \\
 &  & RAG (BM25) & 55.00\,{\scriptsize (5.63)} & 56.33\,{\scriptsize (5.61)} & 57.67\,{\scriptsize (5.59)} & 51.00\,{\scriptsize (5.66)} & \textbf{64.00}\,{\scriptsize (5.43)} & 52.78\,{\scriptsize (4.29)} & 54.44\,{\scriptsize (4.28)} & 55.00\,{\scriptsize (7.27)} & 40.56\,{\scriptsize (4.17)} & \textbf{57.78}\,{\scriptsize (5.22)} \\
 &  & Mem0 & 58.67\,{\scriptsize (5.57)} & 58.33\,{\scriptsize (5.58)} & 60.33\,{\scriptsize (5.54)} & \textbf{61.33}\,{\scriptsize (5.51)} & 58.67\,{\scriptsize (5.57)} & 55.83\,{\scriptsize (5.89)} & 51.67\,{\scriptsize (5.94)} & 56.67\,{\scriptsize (5.87)} & 56.67\,{\scriptsize (5.87)} & \textbf{59.17}\,{\scriptsize (7.79)} \\
\cmidrule(lr){2-13}
 & \multirow{2}{*}{\textbf{Tree-Based}} & MemTree & 57.33\,{\scriptsize (5.60)} & 57.67\,{\scriptsize (5.59)} & 59.00\,{\scriptsize (5.57)} & 43.33\,{\scriptsize (5.61)} & \textbf{66.33}\,{\scriptsize (5.35)} & 56.25\,{\scriptsize (6.28)} & 55.83\,{\scriptsize (6.28)} & \textbf{56.67}\,{\scriptsize (6.27)} & 35.00\,{\scriptsize (6.03)} & 55.83\,{\scriptsize (5.28)} \\
 &  & RAPTOR & 59.33\,{\scriptsize (5.56)} & 57.67\,{\scriptsize (5.59)} & 60.56\,{\scriptsize (4.14)} & 44.67\,{\scriptsize (5.63)} & \textbf{65.67}\,{\scriptsize (5.37)} & 50.33\,{\scriptsize (4.20)} & \textbf{57.78}\,{\scriptsize (5.22)} & 52.22\,{\scriptsize (4.08)} & 37.22\,{\scriptsize (4.06)} & 56.67\,{\scriptsize (5.24)} \\
\cmidrule(lr){2-13}
 &  \multirow{3}{*}{\textbf{Graph-Based}} & Mem0-Graph & 43.33\,{\scriptsize (5.61)} & 43.00\,{\scriptsize (5.60)} & 43.67\,{\scriptsize (5.61)} & 42.67\,{\scriptsize (5.65)} & \textbf{51.00}\,{\scriptsize (5.66)} & 44.17\,{\scriptsize (5.89)} & 48.33\,{\scriptsize (5.94)} & 49.17\,{\scriptsize (5.94)} & 46.67\,{\scriptsize (5.93)} & \textbf{52.16}\,{\scriptsize (5.94)} \\
 &  & A-MEM & 61.67\,{\scriptsize (5.50)} & 61.67\,{\scriptsize (5.50)} & 51.12\,{\scriptsize (5.77)} & 43.67\,{\scriptsize (5.61)} & \textbf{64.00}\,{\scriptsize (5.43)} & 59.17\,{\scriptsize (5.79)} & \textbf{63.33}\,{\scriptsize (5.62)} & 60.00\,{\scriptsize (5.77)} & 37.50\,{\scriptsize (5.66)} & 59.17\,{\scriptsize (5.79)} \\
 &  & GraphRAG & 56.33\,{\scriptsize (5.61)} & 56.67\,{\scriptsize (5.61)} & 57.11\,{\scriptsize (6.14)} & 32.33\,{\scriptsize (5.29)} & \textbf{61.67}\,{\scriptsize (5.50)} & 50.83\,{\scriptsize (5.94)} & 45.83\,{\scriptsize (5.92)} & \textbf{51.67}\,{\scriptsize (5.94)} & 22.50\,{\scriptsize (4.47)} & 50.83\,{\scriptsize (5.94)} \\
\bottomrule
\end{tabular}%
}
\caption{RQ2 comparison of LLM accuracy (\%) across forward reasoning approaches layered on the same memory backbone. Values in parentheses are 95\% confidence-interval half-widths.}
\label{tab:rq2_reasoning_comparison}
\end{table*}

\begin{figure}[!h]
    \centering
    \includegraphics[width=0.95\linewidth]{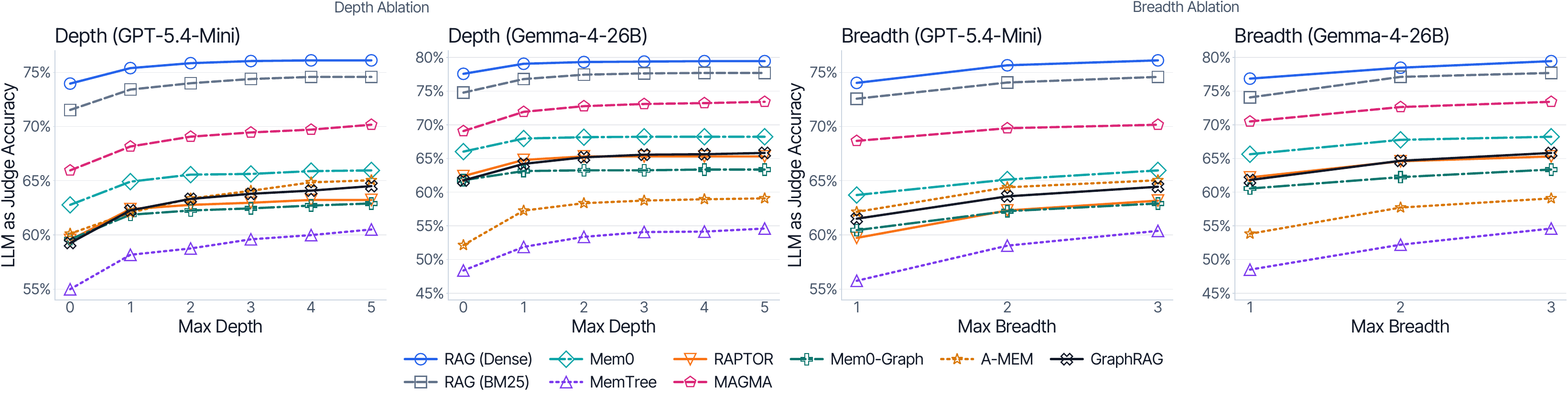}
    \caption{Accuracy vs. $D_{\mathrm{max}}$ (left two) and $B_{\mathrm{max}}$ (right two). Depth yields steady gains, particularly on weaker backbones; breadth saturates after a single decomposition.}
    %Cumulative LLM-as-judge accuracy as a function of maximum depth (left two) and maximum breadth (right two), for \texttt{Gemma-4-26B-A4B} and \texttt{GPT-5.4 mini} backbones. Depth yields steady gains, particularly on weaker backbones; breadth saturates after a single decomposition.}
    \label{fig:ablation}
\end{figure}

\paragraph{RQ3.}
We next ablate the two budgets that govern \textsc{Goal-Mem}'s search: the maximum depth (subgoal refinement) and the maximum breadth (alternative goal decompositions). Figure~\ref{fig:ablation} shows cumulative LLM-as-judge accuracy as each budget is increased, for two backbone LLMs.
Increasing depth yields accuracy improvements until max depth $\geq$ 3, and the gains are largest for backbones with weaker baseline performance. When the first retrieved memory is incomplete or noisy, antecedent-guided refinement lets agents navigate the memory space more systematically, recovering missing context that weaker backbones are especially likely to miss.
Increasing breadth produces little additional gain. We attribute this to the strength of SOTA LLMs; at a single goal, decomposition is typically sufficient, and the depth loop also refines potential failures. Allocating the search budget to depth rather than breadth is therefore the more practical strategy.

\section{Conclusion}
We introduced \textsc{Goal-Mem}, a framework that enhances the reasoning capabilities of memory-augmented LLM agents through goal-oriented backward chaining. By converting user utterances into NL-Logic goals and iteratively grounding subgoals via targeted retrieval and verifiable unification, \textsc{Goal-Mem} enables more effective use of external memory. In addition, its compatibility with diverse memory backbones makes it broadly applicable across existing RAG-based systems. Experimental results show consistent improvements across benchmarks, with particularly strong gains on tasks requiring multi-step inference. Overall, \textsc{Goal-Mem} provides an effective approach for improving memory-grounded reasoning in agentic LLM systems.

\bibliographystyle{plainnat}
\bibliography{reference}
\newpage

\appendix

\section{\textsc{Goal-Mem}  Algorithm}
\label{app:algorithm}
The workflow of the \textsc{Goal-Mem}  reasoning framework is summarized in Algorithm~\ref{alg:goal-mem-patch}. The reasoning process begins by mapping the utterance $u_t$ into a formal goal $\phi_{u_t}$, which specifies the variables that must be instantiated to answer the query. An initially empty substitution $\theta$ is maintained to store variable bindings as they are discovered.

The algorithm proceeds through the \emph{breadth loop}. At each breadth step $b$, the goal $\phi_{u_t}$ is decomposed into a set of subgoals $\psi^{(b)}$ using an LLM-based decomposition function, conditioned on previously explored decompositions $\{\psi^{(0)}, \dots, \psi^{(b-1)}\}$ (with $\psi^{(0)} = \emptyset$). Each subgoal represents a necessary condition for satisfying the original goal.

For each subgoal $\psi_i \in \psi^{(b)}$, the algorithm performs \emph{targeted retrieval} by querying the memory backbone. The retrieved facts are aggregated into a memory set $\mathcal{M}_t^{(b,0)}$. The algorithm then attempts to ground the subgoals via verifiable \emph{unification}, which matches subgoals against retrieved memory while enforcing three constraints: (i) type consistency between variables and candidate values, (ii) consistency with previously established substitutions, and (iii) logical entailment of the grounded subgoal by the retrieved memory (as judged by the LLM). 

This step returns an updated substitution $\theta^{(b,0)}$, a set of unresolved subgoals $\mathcal{U}^{(b,0)}$, and a set of supporting memory facts $\mathcal{F}$ that successfully grounded subgoals. We also maintain the accumulated subgoal set $\Psi^{(b,\le 0)} = \psi^{(b)}$, and define the set of resolved subgoals $\mathcal{S}^{(b,0)} = \Psi^{(b,\le 0)} \setminus \mathcal{U}^{(b,0)}$.

If all subgoals are successfully grounded (i.e., $\mathcal{U}^{(b,0)} = \emptyset$), the algorithm terminates and generates the final response conditioned on the utterance, the substitution, and only the supporting memory $\mathcal{F}$.

Otherwise, the algorithm enters a \emph{depth loop}, which performs recursive refinement of unresolved subgoals. At each depth step $d$, only the subgoals from the previous refinement frontier $\psi^{(b,d-1)}$ (initialized as $\mathcal{U}^{(b,0)}$) are refined using an LLM-based refinement function, producing a new set $\psi^{(b,d)}$ of intermediate subgoals. These refined subgoals are used for additional targeted retrieval, and the newly retrieved facts are accumulated into the memory $\mathcal{M}_t^{(b,d)}$.

The accumulated subgoal set is updated as $\Psi^{(b,\le d)} = \Psi^{(b,\le d-1)} \cup \psi^{(b,d)}$. The unification procedure is then re-applied over $\Psi^{(b,\le d)}$ and $\mathcal{M}_t^{(b,d)}$ to attempt further grounding. To avoid redundant computation while preserving correctness, subgoals that have already been successfully grounded (tracked in $\mathcal{S}^{(b,d-1)}$) are skipped during unification, although they remain part of the accumulated subgoal set for consistency. This step produces updated substitution $\theta^{(b,d)}$, unresolved subgoals $\mathcal{U}^{(b,d)}$, and supporting facts $\mathcal{F}$, and updates the resolved set as $\mathcal{S}^{(b,d)} = \Psi^{(b,\le d)} \setminus \mathcal{U}^{(b,d)}$.

This refinement process continues until either (i) all subgoals are grounded, (ii) no new memory can be retrieved, or (iii) the maximum depth is reached. If grounding succeeds at any depth, the algorithm generates the final response using the grounded substitution and the supporting memory $\mathcal{F}$.

If the current decomposition fails to yield a complete grounding after exhausting the depth budget, the algorithm returns to the breadth loop and considers an alternative decomposition of the original goal, enabling exploration of multiple reasoning strategies.

\begin{algorithm}[t]
\caption{GOAL-MEM: Goal-Oriented Backward Reasoning}
\label{alg:goal-mem-patch}
\begin{algorithmic}[1]
\Require Utterance $u_t$, memory $\mathcal{M}$, retrieval $B_{\text{retrieve}}$, 
max breadth $B_{\mathrm{max}}$, max depth $D_{\mathrm{max}}$
\Ensure Response $a_t$

\State $\phi_{u_t} \gets \textsc{ParseToNLLogic}(u_t)$
\State $\theta^{(0,0)} \gets \emptyset$
\State $\psi^{(0)} \gets \emptyset$

\For{$b = 1$ to $B_{\mathrm{max}}$}

    \State $\psi^{(b)} \gets 
    \Psi^{\text{decomp}}_{\text{LLM}}(\phi_{u_t}, \{\psi^{(0)}, \dots, \psi^{(b-1)}\})$

    \State $\mathcal{M}_t^{(b,0)} \gets \emptyset$
    \ForAll{$\psi_i \in \psi^{(b)}$}
        \State $\mathcal{M}_t^{(b,0)} \gets \mathcal{M}_t^{(b,0)} \cup B_{\text{retrieve}}(\psi_i, M)$
    \EndFor

    \State $(\theta^{(b,0)}, \mathcal{U}^{(b,0)}, \mathcal{F}) \gets 
    \textsc{Unify}(\psi^{(b)}, \mathcal{M}_t^{(b,0)}, \theta^{(0,0)}, \emptyset)$

    \State $\Psi^{(b,\le 0)} \gets \psi^{(b)}$
    \State $\mathcal{S}^{(b,0)} \gets \Psi^{(b,\le 0)} \setminus \mathcal{U}^{(b,0)}$

    \If{$\mathcal{U}^{(b,0)} = \emptyset$}
        \State \Return $\textsc{Generate}(u_t, \theta^{(b,0)}, \mathcal{F})$
    \EndIf

    \State $\psi^{(b,0)} \gets \mathcal{U}^{(b,0)}$

    \For{$d = 1$ to $D_{\mathrm{max}}$}

        \State $\psi^{(b,d)} \gets \emptyset$
        \ForAll{$\psi_i \in \psi^{(b,d-1)}$}
            \State $\psi_i' \gets 
            \Psi^{\text{refine}}_{\text{LLM}}(\psi_i, \mathcal{M}_t^{(b,d-1)})$
            \State $\psi^{(b,d)} \gets \psi^{(b,d)} \cup \{\psi_i'\}$
        \EndFor

        \State $\mathcal{M}_t^{(b,d)} \gets \mathcal{M}_t^{(b,d-1)}$
        \ForAll{$\psi_i' \in \psi^{(b,d)}$}
            \State $\mathcal{M}_t^{(b,d)} \gets \mathcal{M}_t^{(b,d)} \cup B_{\text{retrieve}}(\psi_i', \mathcal{M})$
        \EndFor

        \State $\Psi^{(b,\le d)} \gets \Psi^{(b,\le d-1)} \cup \psi^{(b,d)}$

        \State $(\theta^{(b,d)}, \mathcal{U}^{(b,d)}, \mathcal{F}) \gets
        \textsc{Unify}(\Psi^{(b,\le d)}, \mathcal{M}_t^{(b,d)}, \theta^{(b,d-1)}, \mathcal{S}^{(b,d-1)})$

        \State $\mathcal{S}^{(b,d)} \gets \Psi^{(b,\le d)} \setminus \mathcal{U}^{(b,d)}$

        \If{$\mathcal{U}^{(b,d)} = \emptyset$}
            \State \Return $\textsc{Generate}(u_t, \theta^{(b,d)}, \mathcal{F})$
        \EndIf

        \If{no new memory retrieved}
            \State \textbf{break}
        \EndIf

    \EndFor

\EndFor

\State \Return $\textsc{Failure Flag}$
\end{algorithmic}
\end{algorithm}

\section{Complexity}
\label{appendix:complexity}
Let $K_{\mathrm{max}}$ be an upper bound on the number of subgoals at any depth step. We exclude memory backbone LLM calls, since the latter depend on the implementation of $\mathcal{B}_{\mathrm{retrieve}}$. For each breadth attempt, \textsc{Goal-Mem} uses one LLM call for goal decomposition and one batched LLM call for verifiable unification over all retrieved facts and active subgoals. At each refinement depth, it then uses one LLM call to refine the unresolved subgoals and one batched LLM call to re-run unification after the additional retrieval. Thus, the LLM calls is bounded by $1 + B_{\mathrm{max}}\bigl(2 + 2D_{\mathrm{max}}\bigr)$ or $\mathcal{O}(B_{\mathrm{max}}D_{\mathrm{max}})$ LLM calls. 
The memory retrieval is bounded by:
$
C_{\mathrm{ret}} \leq B_{\mathrm{max}}(D_{\mathrm{max}}+1)K_{\mathrm{max}}.
$

We conduct various empirical analysis in \autoref{fig:ablation} and next paragraph. In practice, the realized cost is often lower because it frequently terminates before the algorithm reaches $B_{\mathrm{max}}$ and $D_{\mathrm{max}}$ (i.e., 95 percentiles of samples have finished within $B = 1$ and $D = 3$ across datasets), as soon as all variables in $\phi_{u_t}$ are grounded. In addition, under batched unification, \textsc{Goal-Mem} uses only a constant number of LLM calls per depth step. Thus, since $B_{\mathrm{max}}$ and $D_{\mathrm{max}}$ are small, \textsc{Goal-Mem}  has a similar asymptotic per-turn LLM-call complexity as existing forward-reasoning approaches.

\subsection{Distribution of Breadth, Depth, and Subgoals in Real Datasets}

\begin{figure}[!h]
    \centering
    \includegraphics[width=0.95\linewidth]{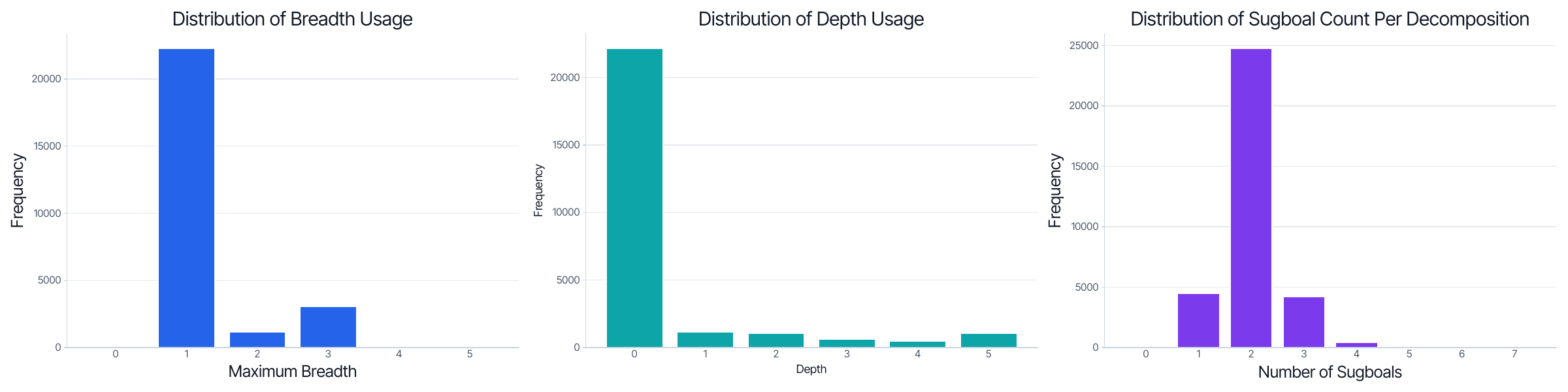}
    \caption{Empirical distributions of realized search statistics in \textsc{Goal-Mem} . The left panel shows the realized breadth usage, the middle panel shows the realized refinement depth, and the right panel shows the number of subgoals generated per decomposition. Although the search caps are $B_{\mathrm{max}}=3$ and $D_{\mathrm{max}}=5$, most examples terminate using substantially smaller realized breadth and depth.}
    \label{fig:realized-search-distribution}
\end{figure}

\autoref{fig:realized-search-distribution} shows that the empirical search cost is concentrated near the smallest possible values. The realized breadth distribution is dominated by breadth $1$, indicating that the first generated goal decomposition is sufficient for most examples. Only a relatively small fraction of cases require additional breadth attempts, and the configured cap $B_{\mathrm{max}}=3$ is reached mainly by harder examples for which the first decomposition does not unify with the retrieved memories.

The realized depth distribution is even more concentrated. Most examples terminate at depth $0$, meaning that \textsc{Goal-Mem}  answers after the initial decomposition, retrieval, and verifiable unification step, without invoking the missing-information refinement loop. A smaller tail reaches depths $1$ through $5$, showing that depth search is used selectively for under-specified or partially grounded questions. Thus, while $D_{\mathrm{max}}=5$ provides a safety cap for difficult cases, the typical realized depth is far below this maximum.

The subgoal-count distribution further supports the efficiency of the method. Most decompositions contain only a small number of subgoals, with the mass concentrated around two subgoals and only a small tail with more complex decompositions. Since memory retrieval scales with the number of queried subgoals, this concentration keeps the realized retrieval fan-out small. Moreover, because unification is batched over the active subgoals and retrieved facts, additional subgoals do not require separate unification LLM calls within the same depth step.

Overall, these distributions show that the theoretical $\mathcal{O}(B_{\mathrm{max}}D_{\mathrm{max}})$ LLM-call bound is a worst-case guarantee rather than the typical operating regime. With $B_{\mathrm{max}}=3$ and $D_{\mathrm{max}}=5$, the search space is tightly bounded, but the realized breadth and depth are usually much smaller. This supports the claim that \textsc{Goal-Mem}  preserves practical per-turn efficiency while still allowing additional breadth and depth when the initial decomposition is insufficient.

\subsection{Token Cost and Wall-Clock Time Across Reasoning Baselines}\label{appendix:implemenntation}

All experiments were run using the OpenAI API (GPT-5.4-mini) and DeepInfra API (Gemma-4-26B-A4B). No local GPU resources were required for backbone LLM inference. Retrieval-based backbones (BM25, Dense) were run on a single CPU machine; graph- and tree-based backbones (Mem0, MAGMA, MemTree, RAPTOR, A-MEM, GraphRAG) were run on a single machine with 16 GB RAM.

\autoref{tab:reasoning_cost_bm25} report the average token cost and wall-clock time across the evaluated forward-reasoning methods using the RAG-BM25 backend. We use RAG-BM25 because it retrieves directly from raw utterances and therefore provides the most consistent token accounting, without additional token overhead introduced by different memory structures. We exclude fixed prompt tokens, since they vary across datasets and use cases and are negligible relative to the retrieved memory facts. We also exclude token usage from memory updating and encoding stages, as these costs depend on the chosen memory backbone and are outside the scope of this work.

Backbone Alone and Query Reformulation incur minimal token cost because they do not require an LLM agent to consume all retrieved memory facts during reasoning. In Query Reformulation, the query is rewritten before the retrieval stage using only the raw user utterance, while Backbone Alone directly retrieves memory facts without an iterative agent loop. In contrast, Self-Reflection, ReAct, MemGuide, and \textsc{Goal-Mem} are iterative memory-conditioned approaches: each step must consume retrieved memory facts to perform reflection, action selection, reranking, or unification for subsequent retrieval. Their token usage is therefore naturally higher than Backbone Alone. However, due to batched unification, \textsc{Goal-Mem} does not repeatedly call the LLM for every subgoal-condition check, allowing it to maintain a competitive token cost compared with other iterative reasoning methods.

\begin{table}[!h]
\centering
\caption{Comparison of runtime and estimated token cost across different reasoning methods with the bm25 backend.}
\label{tab:reasoning_cost_bm25}
\begin{tabular}{lcc}
\toprule
    Method & Avg. End to End Runtime (s) & Avg. Tokens \\
\midrule
Backbone Alone       & 0.68  & 0     \\
Query Reformulation  & 7.52  & 1827  \\
Self-Reflection      & 15.18  & 12269  \\
ReAct                & 16.01  & 22802 \\
\textsc{Goal-Mem}    & 15.10 & 16977  \\
MemGuide             & 14.90 & 14225 \\
\bottomrule
\end{tabular}
\end{table}

\section{Additional Information on the Evaluation Datasets}
\label{appendix:datasets}

We evaluate \textsc{Goal-Mem} on two long-term conversational memory benchmarks: LoCoMo~\citep{maharana2024evaluating} and LongMemEval-Small~\citep{wu2024longmemeval}. These datasets are designed to assess the ability of LLM-based agents to store, retrieve, and reason over information accumulated across extended interactions.

\paragraph{LoCoMo.}
LoCoMo is a benchmark for evaluating long-range conversational memory in multi-session settings. It consists of dialogues that span multiple sessions, where relevant information required to answer a query may be distributed across distant parts of the interaction history. The dataset emphasizes questions that require \emph{multi-hop reasoning}, where the model must retrieve and combine multiple pieces of evidence to arrive at the correct answer. In addition to multi-hop questions, \textsc{LoCoMo} includes single-hop, temporal, and open-domain queries, enabling a fine-grained analysis of model performance across different reasoning types. A key challenge in \textsc{LoCoMo} is that the information necessary to answer a question is often not semantically similar to the query itself, requiring the model to identify and retrieve intermediate facts. This makes it particularly suitable for evaluating the effectiveness of goal-directed retrieval and backward chaining.

\paragraph{\textsc{LongMemEval-Small}.}
\textsc{LongMemEval-Small} is a benchmark focused on core memory capabilities of conversational agents. It evaluates a range of skills, including information extraction from dialogue, reasoning across multiple sessions, handling temporal dependencies, updating knowledge over time, and abstaining when sufficient evidence is not available. Compared to \textsc{LoCoMo}, this dataset places greater emphasis on the correctness and consistency of memory usage, as well as the ability to track evolving user information. The dataset includes diverse query types that test whether the model can accurately retrieve and ground responses in previously stored memory, rather than relying on parametric knowledge.

\paragraph{Evaluation Protocol.}
For both datasets, we evaluate model outputs using an LLM-as-a-judge framework, which assesses the correctness and grounding of generated responses with respect to the reference answers. In addition, we report token-level F1 scores to provide a complementary quantitative measure of answer overlap. These evaluation metrics allow us to capture both semantic correctness and lexical fidelity, offering a comprehensive view of model performance on long-term memory reasoning tasks.

\section{LoCoMo Performance by Question Type}
\label{app:locomo_by_type}
\providecolor{gormblue}{RGB}{235,243,255}
\begin{table*}[t]
\centering
\scriptsize
\setlength{\tabcolsep}{3.0pt}
\renewcommand{\arraystretch}{1.12}
\resizebox{\linewidth}{!}{%
\begin{tabular}{lll c>{\columncolor{gormblue}}c>{\columncolor{gormblue}}c c>{\columncolor{gormblue}}c>{\columncolor{gormblue}}c c>{\columncolor{gormblue}}c>{\columncolor{gormblue}}c c>{\columncolor{gormblue}}c>{\columncolor{gormblue}}c}
\toprule
\multirow{2}{*}{\textbf{LLM}}
& \multirow{2}{*}{\textbf{Category}}
& \multirow{2}{*}{\textbf{Backbone}}
& \multicolumn{3}{c}{\textbf{Multi-hop}\,(\%)}
& \multicolumn{3}{c}{\textbf{Single-hop}\,(\%)}
& \multicolumn{3}{c}{\textbf{Temporal}\,(\%)}
& \multicolumn{3}{c}{\textbf{Open-domain}\,(\%)} \\
\cmidrule(lr){4-6} \cmidrule(lr){7-9} \cmidrule(lr){10-12} \cmidrule(lr){13-15}
& &
& \textbf{Alone} & \textbf{\textsc{Goal-Mem}} & \textbf{$\Delta$} & \textbf{Alone} & \textbf{\textsc{Goal-Mem}} & \textbf{$\Delta$} & \textbf{Alone} & \textbf{\textsc{Goal-Mem}} & \textbf{$\Delta$} & \textbf{Alone} & \textbf{\textsc{Goal-Mem}} & \textbf{$\Delta$} \\
\midrule
\multirow{9}{*}{\shortstack{Gemma\\26B}} & \multirow{3}{*}{\textbf{No Structure}} & RAG (Dense) & 44.33\,{\scriptsize (5.80)} & 63.12\,{\scriptsize (5.63)} & {\footnotesize $\uparrow$18.79} & 48.60\,{\scriptsize (5.47)} & 83.80\,{\scriptsize (4.03)} & {\footnotesize $\uparrow$35.20} & 22.92\,{\scriptsize (8.41)} & 48.96\,{\scriptsize (10.00)} & {\footnotesize $\uparrow$26.04} & 78.12\,{\scriptsize (2.79)} & 86.68\,{\scriptsize (2.30)} & {\footnotesize $\uparrow$8.56} \\
 &  & RAG (BM25) & 37.59\,{\scriptsize (5.65)} & 60.28\,{\scriptsize (5.71)} & {\footnotesize $\uparrow$22.70} & 49.84\,{\scriptsize (5.47)} & 81.62\,{\scriptsize (4.24)} & {\footnotesize $\uparrow$31.78} & 19.79\,{\scriptsize (7.97)} & 48.96\,{\scriptsize (10.00)} & {\footnotesize $\uparrow$29.17} & 76.81\,{\scriptsize (2.85)} & 85.26\,{\scriptsize (2.40)} & {\footnotesize $\uparrow$8.44} \\
 &  & Mem0 & 49.65\,{\scriptsize (5.84)} & 67.38\,{\scriptsize (5.47)} & {\footnotesize $\uparrow$17.73} & 53.27\,{\scriptsize (5.46)} & 56.07\,{\scriptsize (5.43)} & {\footnotesize $\uparrow$2.80} & 42.71\,{\scriptsize (9.90)} & 51.04\,{\scriptsize (10.00)} & {\footnotesize $\uparrow$8.33} & 71.46\,{\scriptsize (3.05)} & 75.03\,{\scriptsize (2.93)} & {\footnotesize $\uparrow$3.57} \\
\midrule
 & \multirow{2}{*}{\textbf{Tree-Based}} & MemTree & 26.24\,{\scriptsize (5.13)} & 37.23\,{\scriptsize (5.64)} & {\footnotesize $\uparrow$10.99} & 32.40\,{\scriptsize (5.12)} & 64.49\,{\scriptsize (5.24)} & {\footnotesize $\uparrow$32.09} & 19.79\,{\scriptsize (7.97)} & 41.67\,{\scriptsize (9.86)} & {\footnotesize $\uparrow$21.88} & 54.10\,{\scriptsize (3.37)} & 58.03\,{\scriptsize (3.34)} & {\footnotesize $\uparrow$3.92} \\
 &  & RAPTOR & 33.69\,{\scriptsize (5.52)} & 61.70\,{\scriptsize (5.67)} & {\footnotesize $\uparrow$28.01} & 32.40\,{\scriptsize (5.12)} & 42.06\,{\scriptsize (5.40)} & {\footnotesize $\uparrow$9.66} & 20.83\,{\scriptsize (8.12)} & 46.88\,{\scriptsize (9.98)} & {\footnotesize $\uparrow$26.04} & 67.42\,{\scriptsize (3.17)} & 77.41\,{\scriptsize (2.83)} & {\footnotesize $\uparrow$9.99} \\
\midrule
 & \multirow{4}{*}{\textbf{Graph-Based}} & MAGMA & 23.40\,{\scriptsize (4.94)} & 51.42\,{\scriptsize (5.83)} & {\footnotesize $\uparrow$28.01} & 58.88\,{\scriptsize (5.38)} & 76.32\,{\scriptsize (4.65)} & {\footnotesize $\uparrow$17.45} & 13.54\,{\scriptsize (6.84)} & 52.08\,{\scriptsize (9.99)} & {\footnotesize $\uparrow$38.54} & 63.73\,{\scriptsize (3.25)} & 82.05\,{\scriptsize (2.59)} & {\footnotesize $\uparrow$18.31} \\
 &  & Mem0-Graph & 51.74\,{\scriptsize (7.47)} & 59.30\,{\scriptsize (7.34)} & {\footnotesize $\uparrow$7.56} & 53.33\,{\scriptsize (7.29)} & 61.67\,{\scriptsize (7.10)} & {\footnotesize $\uparrow$8.33} & 26.42\,{\scriptsize (11.87)} & 33.96\,{\scriptsize (12.75)} & {\footnotesize $\uparrow$7.55} & 68.12\,{\scriptsize (4.17)} & 68.54\,{\scriptsize (4.15)} & {\footnotesize $\uparrow$0.42} \\
 &  & A-MEM & 27.66\,{\scriptsize (5.22)} & 47.87\,{\scriptsize (5.83)} & {\footnotesize $\uparrow$20.21} & 40.50\,{\scriptsize (5.37)} & 63.24\,{\scriptsize (5.27)} & {\footnotesize $\uparrow$22.74} & 18.75\,{\scriptsize (7.81)} & 39.58\,{\scriptsize (9.78)} & {\footnotesize $\uparrow$20.83} & 54.46\,{\scriptsize (3.37)} & 63.50\,{\scriptsize (3.25)} & {\footnotesize $\uparrow$9.04} \\
 &  & GraphRAG & 40.43\,{\scriptsize (5.73)} & 56.38\,{\scriptsize (5.79)} & {\footnotesize $\uparrow$15.96} & 38.63\,{\scriptsize (5.33)} & 44.86\,{\scriptsize (5.44)} & {\footnotesize $\uparrow$6.23} & 21.88\,{\scriptsize (8.27)} & 43.75\,{\scriptsize (9.92)} & {\footnotesize $\uparrow$21.88} & 70.75\,{\scriptsize (3.07)} & 79.43\,{\scriptsize (2.73)} & {\footnotesize $\uparrow$8.68} \\
\midrule
\multirow{9}{*}{\shortstack{GPT-5.4\\mini}} & \multirow{3}{*}{\textbf{No Structure}} & RAG (Dense) & 38.30\,{\scriptsize (5.67)} & 58.51\,{\scriptsize (5.75)} & {\footnotesize $\uparrow$20.21} & 64.49\,{\scriptsize (5.24)} & 81.31\,{\scriptsize (4.26)} & {\footnotesize $\uparrow$16.82} & 44.79\,{\scriptsize (9.95)} & 48.96\,{\scriptsize (10.00)} & {\footnotesize $\uparrow$4.17} & 72.89\,{\scriptsize (3.00)} & 83.08\,{\scriptsize (2.54)} & {\footnotesize $\uparrow$10.19} \\
 &  & RAG (BM25) & 35.11\,{\scriptsize (5.57)} & 53.90\,{\scriptsize (5.82)} & {\footnotesize $\uparrow$18.79} & 73.83\,{\scriptsize (4.81)} & 78.19\,{\scriptsize (4.52)} & {\footnotesize $\uparrow$4.36} & 47.92\,{\scriptsize (9.99)} & 47.92\,{\scriptsize (9.99)} & {\footnotesize $\uparrow$0.00} & 76.93\,{\scriptsize (2.85)} & 83.12\,{\scriptsize (2.53)} & {\footnotesize $\uparrow$6.18} \\
 &  & Mem0 & 48.23\,{\scriptsize (5.83)} & 57.80\,{\scriptsize (5.76)} & {\footnotesize $\uparrow$9.57} & 56.70\,{\scriptsize (5.42)} & 55.14\,{\scriptsize (5.44)} & {\footnotesize $\downarrow$1.56} & 52.08\,{\scriptsize (9.99)} & 50.00\,{\scriptsize (10.00)} & {\footnotesize $\downarrow$2.08} & 72.29\,{\scriptsize (3.02)} & 74.55\,{\scriptsize (2.94)} & {\footnotesize $\uparrow$2.26} \\
\midrule
 & \multirow{2}{*}{\textbf{Tree-Based}} & MemTree & 24.47\,{\scriptsize (5.02)} & 37.59\,{\scriptsize (5.65)} & {\footnotesize $\uparrow$13.12} & 33.96\,{\scriptsize (5.18)} & 65.73\,{\scriptsize (5.19)} & {\footnotesize $\uparrow$31.78} & 32.29\,{\scriptsize (9.35)} & 40.62\,{\scriptsize (9.82)} & {\footnotesize $\uparrow$8.33} & 56.72\,{\scriptsize (3.35)} & 68.37\,{\scriptsize (3.14)} & {\footnotesize $\uparrow$11.65} \\
 &  & RAPTOR & 38.30\,{\scriptsize (5.67)} & 53.90\,{\scriptsize (5.82)} & {\footnotesize $\uparrow$15.60} & 36.76\,{\scriptsize (5.27)} & 37.07\,{\scriptsize (5.28)} & {\footnotesize $\uparrow$0.31} & 42.71\,{\scriptsize (9.90)} & 36.46\,{\scriptsize (9.63)} & {\footnotesize $\downarrow$6.25} & 70.51\,{\scriptsize (3.08)} & 79.31\,{\scriptsize (2.74)} & {\footnotesize $\uparrow$8.80} \\
\midrule
 & \multirow{4}{*}{\textbf{Graph-Based}} & MAGMA & 26.60\,{\scriptsize (5.16)} & 48.94\,{\scriptsize (5.83)} & {\footnotesize $\uparrow$22.34} & 61.06\,{\scriptsize (5.33)} & 74.45\,{\scriptsize (4.77)} & {\footnotesize $\uparrow$13.40} & 41.67\,{\scriptsize (9.86)} & 46.88\,{\scriptsize (9.98)} & {\footnotesize $\uparrow$5.21} & 65.87\,{\scriptsize (3.20)} & 78.24\,{\scriptsize (2.79)} & {\footnotesize $\uparrow$12.37} \\
 &  & Mem0-Graph & 49.65\,{\scriptsize (5.84)} & 57.09\,{\scriptsize (5.78)} & {\footnotesize $\uparrow$7.45} & 52.65\,{\scriptsize (5.46)} & 51.40\,{\scriptsize (5.47)} & {\footnotesize $\downarrow$1.25} & 53.12\,{\scriptsize (9.98)} & 51.04\,{\scriptsize (10.00)} & {\footnotesize $\downarrow$2.08} & 70.51\,{\scriptsize (3.08)} & 70.51\,{\scriptsize (3.08)} & {\footnotesize $\uparrow$0.00} \\
 &  & A-MEM & 24.47\,{\scriptsize (5.02)} & 47.87\,{\scriptsize (5.83)} & {\footnotesize $\uparrow$23.40} & 46.73\,{\scriptsize (5.46)} & 75.08\,{\scriptsize (4.73)} & {\footnotesize $\uparrow$28.35} & 33.33\,{\scriptsize (9.43)} & 45.83\,{\scriptsize (9.97)} & {\footnotesize $\uparrow$12.50} & 46.61\,{\scriptsize (3.37)} & 69.08\,{\scriptsize (3.12)} & {\footnotesize $\uparrow$22.47} \\
 &  & GraphRAG & 37.59\,{\scriptsize (5.65)} & 50.00\,{\scriptsize (5.53)} & {\footnotesize $\uparrow$12.41} & 41.12\,{\scriptsize (5.38)} & 44.69\,{\scriptsize (5.15)} & {\footnotesize $\uparrow$3.57} & 39.58\,{\scriptsize (9.78)} & 47.71\,{\scriptsize (9.38)} & {\footnotesize $\uparrow$8.12} & 73.25\,{\scriptsize (2.99)} & 79.14\,{\scriptsize (2.64)} & {\footnotesize $\uparrow$5.90} \\
\bottomrule
\end{tabular}%
}
\caption{RQ1 breakdown of LLM-as-judge accuracy (\%) by question type (LoCoMo). For each type, we report the backbone-alone score, the score after augmenting the same backbone with \textsc{Goal-Mem}, and the absolute lift ($\Delta$). Adversarial questions are omitted. Values in parentheses are 95\% confidence-interval half-widths .}
\label{tab:rq1_question_type_breakdown}
\end{table*}

\providecolor{gormblue}{RGB}{235,243,255}
\begin{table*}[t]
\centering
\scriptsize
\setlength{\tabcolsep}{3.0pt}
\renewcommand{\arraystretch}{1.12}
\resizebox{\linewidth}{!}{%
\begin{tabular}{lll c>{\columncolor{gormblue}}c>{\columncolor{gormblue}}c c>{\columncolor{gormblue}}c>{\columncolor{gormblue}}c c>{\columncolor{gormblue}}c>{\columncolor{gormblue}}c c>{\columncolor{gormblue}}c>{\columncolor{gormblue}}c}
\toprule
\multirow{2}{*}{\textbf{LLM}}
& \multirow{2}{*}{\textbf{Category}}
& \multirow{2}{*}{\textbf{Backbone}}
& \multicolumn{3}{c}{\textbf{Multi-hop}}
& \multicolumn{3}{c}{\textbf{Single-hop}}
& \multicolumn{3}{c}{\textbf{Temporal}}
& \multicolumn{3}{c}{\textbf{Open-domain}} \\
\cmidrule(lr){4-6} \cmidrule(lr){7-9} \cmidrule(lr){10-12} \cmidrule(lr){13-15}
& &
& \textbf{Alone} & \textbf{\textsc{Goal-Mem}} & \textbf{$\Delta$} & \textbf{Alone} & \textbf{\textsc{Goal-Mem}} & \textbf{$\Delta$} & \textbf{Alone} & \textbf{\textsc{Goal-Mem}} & \textbf{$\Delta$} & \textbf{Alone} & \textbf{\textsc{Goal-Mem}} & \textbf{$\Delta$} \\
\midrule
\multirow{9}{*}{\shortstack{Gemma\\26B}} & \multirow{3}{*}{\textbf{No Structure}} & RAG (Dense) & 0.390\,{\scriptsize (0.038)} & 0.501\,{\scriptsize (0.039)} & $\uparrow$0.111 & 0.458\,{\scriptsize (0.030)} & 0.644\,{\scriptsize (0.037)} & $\uparrow$0.187 & 0.145\,{\scriptsize (0.059)} & 0.339\,{\scriptsize (0.084)} & $\uparrow$0.194 & 0.674\,{\scriptsize (0.024)} & 0.680\,{\scriptsize (0.024)} & $\uparrow$0.007 \\
 &  & RAG (BM25) & 0.341\,{\scriptsize (0.037)} & 0.472\,{\scriptsize (0.039)} & $\uparrow$0.131 & 0.451\,{\scriptsize (0.032)} & 0.637\,{\scriptsize (0.037)} & $\uparrow$0.186 & 0.120\,{\scriptsize (0.055)} & 0.323\,{\scriptsize (0.084)} & $\uparrow$0.203 & 0.671\,{\scriptsize (0.024)} & 0.681\,{\scriptsize (0.024)} & $\uparrow$0.010 \\
 &  & Mem0 & 0.398\,{\scriptsize (0.038)} & 0.479\,{\scriptsize (0.039)} & $\uparrow$0.081 & 0.495\,{\scriptsize (0.037)} & 0.543\,{\scriptsize (0.038)} & $\uparrow$0.048 & 0.223\,{\scriptsize (0.065)} & 0.343\,{\scriptsize (0.085)} & $\uparrow$0.120 & 0.540\,{\scriptsize (0.027)} & 0.524\,{\scriptsize (0.026)} & $\downarrow$0.015 \\
\midrule
 & \multirow{2}{*}{\textbf{Tree-Based}} & MemTree & 0.207\,{\scriptsize (0.032)} & 0.314\,{\scriptsize (0.039)} & $\uparrow$0.107 & 0.286\,{\scriptsize (0.034)} & 0.514\,{\scriptsize (0.043)} & $\uparrow$0.228 & 0.064\,{\scriptsize (0.029)} & 0.285\,{\scriptsize (0.082)} & $\uparrow$0.221 & 0.423\,{\scriptsize (0.027)} & 0.425\,{\scriptsize (0.028)} & $\uparrow$0.002 \\
 &  & RAPTOR & 0.329\,{\scriptsize (0.037)} & 0.482\,{\scriptsize (0.040)} & $\uparrow$0.153 & 0.287\,{\scriptsize (0.029)} & 0.410\,{\scriptsize (0.040)} & $\uparrow$0.123 & 0.122\,{\scriptsize (0.054)} & 0.284\,{\scriptsize (0.082)} & $\uparrow$0.162 & 0.576\,{\scriptsize (0.027)} & 0.601\,{\scriptsize (0.026)} & $\uparrow$0.026 \\
\midrule
 & \multirow{4}{*}{\textbf{Graph-Based}} & MAGMA & 0.237\,{\scriptsize (0.034)} & 0.412\,{\scriptsize (0.037)} & $\uparrow$0.175 & 0.403\,{\scriptsize (0.037)} & 0.590\,{\scriptsize (0.037)} & $\uparrow$0.187 & 0.087\,{\scriptsize (0.036)} & 0.331\,{\scriptsize (0.083)} & $\uparrow$0.243 & 0.547\,{\scriptsize (0.027)} & 0.643\,{\scriptsize (0.025)} & $\uparrow$0.096 \\
 &  & Mem0-Graph & 0.385\,{\scriptsize (0.050)} & 0.450\,{\scriptsize (0.053)} & $\uparrow$0.065 & 0.454\,{\scriptsize (0.048)} & 0.546\,{\scriptsize (0.050)} & $\uparrow$0.092 & 0.160\,{\scriptsize (0.075)} & 0.239\,{\scriptsize (0.102)} & $\uparrow$0.079 & 0.506\,{\scriptsize (0.036)} & 0.473\,{\scriptsize (0.036)} & $\downarrow$0.033 \\
 &  & A-MEM & 0.265\,{\scriptsize (0.034)} & 0.355\,{\scriptsize (0.037)} & $\uparrow$0.091 & 0.363\,{\scriptsize (0.034)} & 0.553\,{\scriptsize (0.041)} & $\uparrow$0.190 & 0.080\,{\scriptsize (0.043)} & 0.269\,{\scriptsize (0.079)} & $\uparrow$0.189 & 0.471\,{\scriptsize (0.028)} & 0.476\,{\scriptsize (0.028)} & $\uparrow$0.005 \\
 &  & GraphRAG & 0.342\,{\scriptsize (0.038)} & 0.451\,{\scriptsize (0.038)} & $\uparrow$0.109 & 0.282\,{\scriptsize (0.032)} & 0.409\,{\scriptsize (0.040)} & $\uparrow$0.127 & 0.120\,{\scriptsize (0.049)} & 0.290\,{\scriptsize (0.082)} & $\uparrow$0.171 & 0.600\,{\scriptsize (0.026)} & 0.596\,{\scriptsize (0.026)} & $\downarrow$0.005 \\
\midrule
\multirow{9}{*}{\shortstack{GPT-5.4\\mini}} & \multirow{3}{*}{\textbf{No Structure}} & RAG (Dense) & 0.344\,{\scriptsize (0.038)} & 0.441\,{\scriptsize (0.039)} & $\uparrow$0.097 & 0.492\,{\scriptsize (0.035)} & 0.583\,{\scriptsize (0.037)} & $\uparrow$0.091 & 0.283\,{\scriptsize (0.077)} & 0.353\,{\scriptsize (0.085)} & $\uparrow$0.070 & 0.616\,{\scriptsize (0.026)} & 0.660\,{\scriptsize (0.025)} & $\uparrow$0.044 \\
 &  & RAG (BM25) & 0.304\,{\scriptsize (0.035)} & 0.423\,{\scriptsize (0.039)} & $\uparrow$0.120 & 0.520\,{\scriptsize (0.035)} & 0.573\,{\scriptsize (0.038)} & $\uparrow$0.053 & 0.273\,{\scriptsize (0.074)} & 0.345\,{\scriptsize (0.085)} & $\uparrow$0.072 & 0.663\,{\scriptsize (0.024)} & 0.648\,{\scriptsize (0.025)} & $\downarrow$0.015 \\
 &  & Mem0 & 0.395\,{\scriptsize (0.038)} & 0.414\,{\scriptsize (0.039)} & $\uparrow$0.019 & 0.497\,{\scriptsize (0.034)} & 0.524\,{\scriptsize (0.036)} & $\uparrow$0.027 & 0.379\,{\scriptsize (0.087)} & 0.330\,{\scriptsize (0.082)} & $\downarrow$0.049 & 0.561\,{\scriptsize (0.026)} & 0.542\,{\scriptsize (0.026)} & $\downarrow$0.019 \\
\midrule
 & \multirow{2}{*}{\textbf{Tree-Based}} & MemTree & 0.232\,{\scriptsize (0.034)} & 0.304\,{\scriptsize (0.039)} & $\uparrow$0.073 & 0.378\,{\scriptsize (0.033)} & 0.522\,{\scriptsize (0.039)} & $\uparrow$0.144 & 0.177\,{\scriptsize (0.065)} & 0.275\,{\scriptsize (0.079)} & $\uparrow$0.098 & 0.469\,{\scriptsize (0.028)} & 0.526\,{\scriptsize (0.028)} & $\uparrow$0.057 \\
 &  & RAPTOR & 0.346\,{\scriptsize (0.037)} & 0.409\,{\scriptsize (0.039)} & $\uparrow$0.063 & 0.358\,{\scriptsize (0.030)} & 0.371\,{\scriptsize (0.034)} & $\uparrow$0.013 & 0.284\,{\scriptsize (0.075)} & 0.232\,{\scriptsize (0.077)} & $\downarrow$0.052 & 0.591\,{\scriptsize (0.026)} & 0.649\,{\scriptsize (0.025)} & $\uparrow$0.058 \\
\midrule
 & \multirow{4}{*}{\textbf{Graph-Based}} & MAGMA & 0.241\,{\scriptsize (0.034)} & 0.390\,{\scriptsize (0.037)} & $\uparrow$0.149 & 0.465\,{\scriptsize (0.035)} & 0.550\,{\scriptsize (0.035)} & $\uparrow$0.085 & 0.290\,{\scriptsize (0.078)} & 0.310\,{\scriptsize (0.082)} & $\uparrow$0.019 & 0.563\,{\scriptsize (0.026)} & 0.631\,{\scriptsize (0.026)} & $\uparrow$0.069 \\
 &  & Mem0-Graph & 0.356\,{\scriptsize (0.038)} & 0.392\,{\scriptsize (0.040)} & $\uparrow$0.036 & 0.465\,{\scriptsize (0.033)} & 0.459\,{\scriptsize (0.036)} & $\downarrow$0.005 & 0.383\,{\scriptsize (0.087)} & 0.335\,{\scriptsize (0.085)} & $\downarrow$0.048 & 0.520\,{\scriptsize (0.027)} & 0.504\,{\scriptsize (0.027)} & $\downarrow$0.016 \\
 &  & A-MEM & 0.247\,{\scriptsize (0.033)} & 0.384\,{\scriptsize (0.038)} & $\uparrow$0.137 & 0.472\,{\scriptsize (0.036)} & 0.580\,{\scriptsize (0.038)} & $\uparrow$0.108 & 0.222\,{\scriptsize (0.071)} & 0.341\,{\scriptsize (0.086)} & $\uparrow$0.119 & 0.404\,{\scriptsize (0.027)} & 0.570\,{\scriptsize (0.027)} & $\uparrow$0.166 \\
 &  & GraphRAG & 0.298\,{\scriptsize (0.036)} & 0.365\,{\scriptsize (0.036)} & $\uparrow$0.067 & 0.356\,{\scriptsize (0.031)} & 0.414\,{\scriptsize (0.032)} & $\uparrow$0.059 & 0.276\,{\scriptsize (0.079)} & 0.326\,{\scriptsize (0.078)} & $\uparrow$0.050 & 0.604\,{\scriptsize (0.026)} & 0.619\,{\scriptsize (0.025)} & $\uparrow$0.015 \\
\bottomrule
\end{tabular}%
}
\caption{RQ1 breakdown of token-level F1 by question type (LoCoMo). For each type, we report the backbone-alone score, the score after augmenting the same backbone with \textsc{Goal-Mem}, and the absolute lift ($\Delta$). Adversarial questions are omitted. Values in parentheses are 95\% confidence-interval half-widths .}
\label{tab:rq1_question_type_breakdown_f1}
\end{table*}

\autoref{tab:rq1_question_type_breakdown} and \autoref{tab:rq1_question_type_breakdown_f1} present detailed performance breakdowns on the LoCoMo dataset by question type. Notably, we observe a consistent improvement on multi-hop questions across both LLM backbones.

% Required packages if not already in the paper preamble: \usepackage{booktabs,tabularx,hyperref}
% This section assumes the forward-baseline method names are defined elsewhere in the appendix.

\section{Qualitative Analysis of Multi-hop Forward-Reasoning Failures}
\label{app:qualitative-multihop-failures}

We select the failure examples in \autoref{tab:qualitative-multihop-failures} from LoCoMo via RAG-BM25 memory backbone since it has the the most comprehensive memory traces.

\begin{table*}[!h]
\centering
\small
\setlength{\tabcolsep}{3pt}
\renewcommand{\arraystretch}{1.15}
\begin{tabularx}{\linewidth}{p{0.15\linewidth}p{0.1\linewidth}p{0.25\linewidth}p{0.40\linewidth}}
\toprule \textbf{Question} & \textbf{Goal-Mem} & \textbf{Forward baseline outputs} & \textbf{Goal-Mem reasoning pattern} \\
\midrule
In which state is the shelter from which James adopted the puppy? & Connecticut & All forward baselines return Stamford. & Grounds the adoption memory to the shelter location Stamford, keeps the requested answer type as a state, and applies the location-granularity bridge Stamford $\rightarrow$ Connecticut. \\
\midrule
 What game was James playing in the online gaming tournament in April 2022? & Apex Legends & Query Reformulation, Self-Reflection, and MemGuide return ``I don't know''; ReAct returns Civilization VI. & Grounds the April tournament event, refines the missing game slot, and links the temporally aligned game mention to Apex Legends rather than a different gaming memory. \\
\midrule
 What is the favorite game Jolene plays with her partner? & It Takes Two & Query Reformulation, Self-Reflection, and ReAct return Detroit; MemGuide returns that it is hard to choose one. & Compares multiple partner-gaming memories and accepts the memory that explicitly entails favorite/shared enjoyment: Jolene says she loves playing It Takes Two together. \\
\midrule
What outdoor activity did Jolene suggest doing together with Deborah? & Surfing & Query Reformulation returns none; Self-Reflection and MemGuide say no outdoor activity was suggested; ReAct returns ``I don't know.'' & Uses a later breadth decomposition that searches for an activity Jolene is learning and wants to do with Deborah; unifies the question with the surfing exchange and the invitation to ``come with me.'' \\
\midrule
What area was hit by a flood? & West County & Query Reformulation, Self-Reflection, and MemGuide return ``my old area''; ReAct says the memory is unavailable. & Treats ``my old area'' as an unresolved referring expression, then grounds it with an earlier memory where John identifies his old area as West County. \\
\midrule
What fun activity did Evan mention doing in July 2023? & Skiing & Query Reformulation and Self-Reflection return ``drove somewhere fun''; ReAct returns a mystery novel; MemGuide returns \emph{The Great Gatsby}. & Uses temporal alignment: a later August memory says Evan drove somewhere fun ``last month,'' and the follow-up memory identifies the activity as skiing, which places the activity in July 2023. \\
\bottomrule
\end{tabularx}
\caption{Qualitative examples where all four forward-reasoning baselines fail but \textsc{Goal-Mem} returns the correct answer. The examples illustrate different multi-hop patterns: location-granularity bridging, temporal event alignment, distractor disambiguation, breadth search over alternative decompositions, and grounding of underspecified referring expressions.}
\label{tab:qualitative-multihop-failures}
\end{table*}

The forward baselines are strong retrieval-augmented methods, but they primarily decide what to retrieve from the surface form of the question and the currently accumulated evidence. As a result, they often retrieve a salient memory and then answer with the most lexical or locally plausible span. For example , the forward methods retrieve the shelter memory but return the city Stamford, even though the question asks for the state. Or several baselines return the unresolved phrase ``my old area,'' which is evidence that the flood memory was found but the required referring-expression grounding was not completed.

\textsc{Goal-Mem} instead of treating retrieved memories as a flat context for generation, it decomposes the question into checkable subgoals and then applies verifiable unification. This allows the model to keep track of answer variables, requested answer types, and unresolved referring expressions. For example, the goal is not merely to find any game-related memory, but to bind the variable corresponding to the game played in James's April online tournament. The first pass does not directly ground this variable, so the missing-information refinement step asks for the specific game associated with that tournament. The later evidence grounds the same event to Apex Legends, while unrelated games such as Civilization VI are rejected because they do not satisfy the tournament-and-time constraint.

These examples also show depth is useful when the current decomposition is correct but one variable remains ungrounded, as in the April tournament example. Breadth is useful when the first decomposition is too literal or retrieves only weak evidence. The system initially fails to find a direct phrasing of ``suggested outdoor activity.'' A later breadth attempt reformulates the goal as evidence that Jolene is interested in an activity and wants to do it with Deborah.

The failure cases highlight that multi-hop memory QA is not only a retrieval-volume problem. The relevant evidence is often present in the retrieved pool, but the final answer requires an additional grounding step: resolving a location to the requested granularity, linking a relative time expression to an event month, choosing the memory that entails the queried relation, or replacing a referring expression with its named antecedent. \textsc{Goal-Mem}'s decomposition and unification process makes these intermediate requirements explicit, which helps prevent the model from returning a retrieved but ungrounded span as the final answer.
\section{Forward Reasoning Baseline Implementations}
\label{app:forward-baselines}

This appendix describes the forward reasoning baselines used in RQ2: Query Reformulation, Self-Reflection, ReAct, and MemGuide. All four methods are implemented as wrappers around the same retrieval-augmented memory backbone $\mathcal{B}$ used by \textsc{Goal-Mem}. Thus, these baselines do not change the memory storage, indexing, or update procedure of the underlying backbone; they only change how queries are generated, how retrieved memory units are accumulated or filtered, and how the final answer is produced.

\paragraph{Shared retrieval protocol.}
For a controlled comparison, all baselines use the same inference LLM, decoding configuration, memory backbone $\mathcal{B}$, and answer-generation prompt as the corresponding \textsc{Goal-Mem} run. We use a shared retrieved-memory cap of
\begin{equation}
    k_{\max}^{\mathrm{ret}} = 60,
\end{equation}
where $k_{\max}^{\mathrm{ret}}$ denotes the maximum number of unique retrieved memory units that can be exposed to the final answer stage for one question. This retrieval cap is distinct from $K_{\max}$ in the complexity analysis, where $K_{\max}$ denotes the maximum number of subgoals produced by a goal decomposition. For iterative baselines, the same cap is distributed across retrieval actions: each retrieval action calls $\mathcal{B}_{\mathrm{retrieve}}$ with a bounded per-query fan-out, retrieved units are deduplicated, and the merged pool is clipped to at most $k_{\max}^{\mathrm{ret}}$ units before final answer generation or filtering.

\paragraph{Query Reformulation.}
The Query Reformulation baseline follows the query-rewriting view of retrieval-augmented generation, where an LLM adapts the user input before retrieval to reduce the gap between the surface form of the question and the information needed by the reader~\citep{ma2023query}. Given the user utterance $u_t$, the LLM generates $n_{\mathrm{qr}}=5$ diverse retrieval queries. The original question is also included as a retrieval query, giving
\begin{equation}
    Q_{\mathrm{qr}}(u_t)
    = \{u_t\} \cup \operatorname{Rewrite}_{\mathrm{LLM}}(u_t; n_{\mathrm{qr}}=3).
\end{equation}
The prompt asks the LLM to target different entities, events, temporal expressions, synonyms, and background aspects that may appear in memory, while preserving important entities and events from the original question. The retrieved results from all queries are merged, deduplicated, and clipped to $k_{\max}^{\mathrm{ret}}=60$. The final answer is then generated using the standard answer prompt over the resulting memory pool. This baseline is intentionally simple: it performs no symbolic decomposition, no unification, no missing-slot refinement, and no post-retrieval reranking.

\paragraph{Self-Reflection.}
The Self-Reflection baseline adapts the reflective-agent principle of Reflexion to the single-question memory retrieval setting. Reflexion uses verbal feedback and an episodic memory buffer to improve later decisions; in our evaluation, we use the same forward-style reflection idea without carrying reflections across evaluation questions, so that the baseline does not receive extra persistent feedback unavailable to the other methods. The method first retrieves memories using the original question. At each reflection turn, the LLM is shown the question, all previously issued queries, and the accumulated retrieved memory pool. It outputs a JSON decision with three fields: whether the pool is sufficient, one concise next query if more evidence is needed, and a short rationale. If the pool is judged sufficient, the loop terminates; otherwise, the follow-up query is sent to $\mathcal{B}_{\mathrm{retrieve}}$, and the newly retrieved units are merged and deduplicated with the existing pool.

We set the maximum number of reflection turns to $T_{\max}^{\mathrm{sr}}=5$. The method can therefore terminate because the LLM judges the memory pool sufficient, because no new retrieval query is proposed, because retrieval returns no new memory units, or because the turn limit is reached. The final answer is generated from the accumulated memory pool after clipping to $k_{\max}^{\mathrm{ret}}=60$. Unlike \textsc{Goal-Mem}, Self-Reflection does not maintain typed variables or substitutions and does not verify that a retrieved memory entails a particular subgoal; each additional query is generated forward from the current question and evidence pool.

\paragraph{ReAct.}
The ReAct baseline follows the reasoning-and-acting paradigm in which the LLM alternates between natural-language reasoning traces and actions that interact with an external environment~\citep{yao2022react}. In our memory-QA adaptation, the external environment is the shared memory backbone $\mathcal{B}$. At each step, the model emits exactly one \texttt{Thought} and one \texttt{Action}. The action space is restricted to
\begin{equation}
    \texttt{Retrieve[query]} \quad \text{or} \quad \texttt{Finish[answer]}.
\end{equation}
When the model selects \texttt{Retrieve[query]}, the query is sent to $\mathcal{B}_{\mathrm{retrieve}}$, and the returned memories are appended to the trajectory as an \texttt{Observation}. When the model selects \texttt{Finish[answer]}, the answer is returned directly. This preserves the ReAct design choice of interleaving reasoning traces with environment actions, while making the environment identical to the memory backend used by \textsc{Goal-Mem}.

We set the maximum number of ReAct steps to $S_{\max}^{\mathrm{react}}=5$. If the model output does not match the required \texttt{Thought}/\texttt{Action} format, the implementation retries once with a stricter format reminder. If the step limit is reached before a \texttt{Finish} action, the model is forced to produce a final \texttt{Finish} answer using only the observations collected so far. Retrieved observations are deduplicated, and the accumulated memory pool is clipped to $k_{\max}^{\mathrm{ret}}=60$. As with Self-Reflection, ReAct is a forward reasoning baseline: it decides the next retrieval action from the current trajectory, rather than decomposing the target answer into subgoals and verifying substitutions.

\paragraph{MemGuide.}
MemGuide is a memory-selection method for goal-oriented multi-session LLM agents. The original method introduces a two-stage intent-driven framework: Intent-Aligned Retrieval retrieves memory units that share the current task intent, and Missing-Slot Guided Filtering reranks candidate memory units according to their marginal gain in filling missing task slots~\citep{du2025memguide}. Our implementation follows this two-stage structure while adapting it to the unified memory-backbone interface used in our experiments.

First, an intent-capture prompt maps the question to a compact intent description and generates $n_{\mathrm{intent}}=5$ intent-aligned retrieval queries. The original question is included as an additional seed query. These queries retrieve an initial candidate pool from $\mathcal{B}$. Second, a missing-slot reasoner inspects the current candidate pool and decides whether the pool is sufficient. If not, it enumerates concrete missing fields, such as a date, location, counterpart person, event outcome, or relation, and proposes at most $F_{\max}^{\mathrm{mg}}=5$ follow-up queries for the next refinement round. We allow at most $R_{\max}^{\mathrm{mg}}=3$ missing-slot refinement rounds.

After retrieval, MemGuide applies slot-guided filtering. Candidate memory units are scored by an LLM judger according to how directly they answer the question or fill the identified missing slots. The implementation combines the LLM score with the retrieval prior,
\begin{equation}
    s(f)
    = \lambda s_{\mathrm{LLM}}(f)
    + (1-\lambda)s_{\mathrm{ret}}(f),
    \qquad \lambda = 0.7,
\end{equation}
where $s_{\mathrm{ret}}(f)$ favors earlier retrieved candidates. The final selected memory pool is sorted by this score and clipped to $k_{\max}^{\mathrm{ret}}=60$ before answer generation. The candidate pool is also clipped to the same cap after retrieval and deduplication.

This adaptation is faithful to the algorithmic role of MemGuide as an intent- and slot-aware forward memory selector, but it does not assume MemGuide's original task-oriented-dialogue memory format or its task-specific fine-tuned filtering model. Instead, all candidate units are retrieved from the same memory backbone $\mathcal{B}$ used by the other baselines, and the slot filter is implemented with the shared inference LLM. This keeps the comparison controlled: MemGuide receives the benefit of intent-aligned retrieval and missing-slot filtering, but it does not receive a method-specific memory representation or extra supervised training signal unavailable to the other methods.

\section{Broader Impacts}\label{appendix:impacts}
Improving the coherence and factual grounding of long-horizon conversational agents has clear positive societal value in applications such as healthcare assistance, education, and personal productivity. By anchoring responses in retrieved memory rather than parametric knowledge, \textsc{Goal-Mem} reduces the risk of hallucinated but plausible-sounding responses. However, as with all existing RAG-based memory works and LLM-based systems, there may be concerns about the possibility of mistakes and the risk of over-reliance on the system's outputs.

\section{Limitations}
\label{appendix:limitations}
As we observed in Section~\ref{sec:results}, \textsc{Goal-Mem} consistently improves the performance of existing RAG-based agentic memory systems. At the same time, since it builds upon an underlying memory and retrieval backbone and leverages an LLM for components such as goal decomposition and answer generation, its effectiveness naturally depends on the quality of these components. %Moreover, as with every other approach for reasoning enhancement of LLM-based agents, \textsc{Goal-Mem} is possible to make errors, and thus, overreliance on its answers must be avoided.

\section{Complete Prompts}
In this appendix, we disclose the prompt we used in this experiment for \textsc{Goal-Mem}.
\subsection{Goal Decomposition}

\subsubsection{Initial decomposition}
\begin{lstlisting}[style=goalmemprompt]
system: |-
  system_commonsense: |-
  You will be provided only with a question about a person.

  Generate one concise commonsense goal decomposition that can guide later evidence retrieval and answer verification.

  Requirements:
  - Do NOT use any specific retrieved facts or invented details.
  - Use general human commonsense only.
  - Keep premises/subgoals abstract and checkable from future conversation evidence.
  - The decomposition should help identify what evidence would be sufficient to answer the question.
  - Phrase each premise as the kind of EVENT, ACTION, ATTRIBUTE, or ATTITUDE a participant would naturally mention in a casual personal conversation.
  - DO NOT phrase premises as META-CLAIMS about the conversation itself. FORBIDDEN openings include: "the conversation explicitly states ...", "the conversation gives ...", "the message names ...", "a fact says ...", "the speaker explicitly mentions ...".
  - Prefer concrete event/attribute language ("the person visits Stamford", "the person talks about a wedding in December") over evidentiary framing ("the conversation contains a statement about a wedding in December").
  - Preserve exact entity anchors from the question and make answer variables explicit as (x:type), (y:type), etc.

  Goal decomposition:
  Step 1 - Identify the central entities in the question: subject, object/topic/instrument/place/activity, named person, time qualifier, and requested answer type.
  Step 2 - Identify any variables that must be substituted to answer the question. Write variables as (x:type), (y:type), etc.
  Step 3 - Generate atomic subgoals. Each subgoal must be factual, verifiable, and phrased as the kind of event, action, attribute, or attitude that could naturally appear in a personal conversation memory.
  Step 4 - Preserve exact entity anchors from the question. Closely related but distinct entities are not substitutes (e.g. guitar vs violin; Korean class vs Korea trip; current role vs previous role).
  Step 5 - Use facts only as optional grounding context. Do NOT answer the question and do NOT invent concrete substitutions.


    Output format (exactly):
  Goal: <one goal for the question, using typed variables where needed>
  Rule: IF <subgoal 1> AND <subgoal 2> AND ... THEN <goal is satisfied>
  Variables:
  - (<variable:type>): <what this variable must be substituted with, or "none" if no answer variable is needed>
  Subgoals:
  - <atomic subgoal 1>
  - <atomic subgoal 2>
  Premises:
  - <same text as subgoal 1, repeated for backward compatibility with retrieval code>
  - <same text as subgoal 2, repeated for backward compatibility with retrieval code>

input_template_commonsense: |-
  Question: {{QUESTION}}
  {% if BACKGROUND_SUMMARY %}
  Background (per-speaker conversation summary):
  {{BACKGROUND_SUMMARY}}
  Note on the background: this is OPTIONAL context to help you ground the
  speakers — it is NOT exhaustive and NOT a fact source. Do NOT restrict your
  rule to only what appears here, do NOT cite the summary as evidence, and do
  NOT treat absence from the summary as absence from the conversation. General
  commonsense reasoning should still be the primary driver of the rule; use the
  background only as light supporting context.
  {% endif %}
  {% if ASSUMED_KNOWN_INFO %}
  Known Info (assume true): {{ASSUMED_KNOWN_INFO|join('\n')}}
  {% endif %}

few-shot:
  - input:
      QUESTION: |-
        What career field is Caroline likely to pursue in her education?
    output: |-
      Goal: (Caroline) is likely to pursue (x:career_field) in her education.
      Rule: IF Caroline is actively exploring (x:career_field) AND Caroline has expressed direct interest in (x:career_field) AND Caroline finds work in (x:career_field) meaningful, THEN Caroline is likely to pursue (x:career_field) in her education.
      Variables:
      - (x:career_field): the education or career field Caroline is likely to pursue
      Subgoals:
      - Caroline is actively exploring (x:career_field).
      - Caroline has expressed direct interest in (x:career_field).
      - Caroline finds work in (x:career_field) meaningful.
      Premises:
      - Caroline is actively exploring (x:career_field).
      - Caroline has expressed direct interest in (x:career_field).
      - Caroline finds work in (x:career_field) meaningful.
  - input:
      QUESTION: |-
        Would Melanie be likely to attend an outdoor fitness event?
    output: |-
      Goal: (Melanie) is likely to attend an outdoor fitness event.
      Rule: IF Melanie regularly engages in outdoor physical activity AND Melanie is training for a competitive running event, THEN Melanie would likely be interested in attending an outdoor fitness event.
      Variables:
      - none: no answer variable is needed because the question asks for a yes/no judgment
      Subgoals:
      - Melanie regularly engages in outdoor physical activity.
      - Melanie is training for a competitive running event.
      Premises:
      - Melanie regularly engages in outdoor physical activity.
      - Melanie is training for a competitive running event.
  - input:
      QUESTION: |-
        Did Caroline use art as a way to explore personal identity?
    output: |-
      Goal: (Caroline) used art as a way to explore personal identity.
      Rule: IF Caroline has used art as a medium to explore identity AND Caroline values art as a meaningful practice, THEN Caroline used art as a way to explore personal identity.
      Variables:
      - none: no answer variable is needed because the question asks for a yes/no judgment
      Subgoals:
      - Caroline has used art as a medium to explore identity.
      - Caroline values art as a meaningful practice.
      Premises:
      - Caroline has used art as a medium to explore identity.
      - Caroline values art as a meaningful practice.
  - input:
      QUESTION: |-
        What career field is Caroline likely to pursue in her education?
    output: |-
      Goal: (Caroline) is likely to pursue (x:career_field) in her education.
      Rule: IF Caroline consistently expresses sustained interest in (x:career_field) AND Caroline takes concrete steps toward training or education in (x:career_field), THEN Caroline is likely to pursue (x:career_field) in her education.
      Variables:
      - (x:career_field): the education or career field Caroline is likely to pursue
      Subgoals:
      - Caroline consistently expresses sustained interest in (x:career_field).
      - Caroline takes concrete steps toward training or education in (x:career_field).
      Premises:
      - Caroline consistently expresses sustained interest in (x:career_field).
      - Caroline takes concrete steps toward training or education in (x:career_field).
  - input:
      QUESTION: |-
        Would Melanie be likely to attend an outdoor fitness event?
    output: |-
      Goal: (Melanie) is likely to attend an outdoor fitness event.
      Rule: IF Melanie regularly engages in outdoor exercise AND Melanie shows motivation for fitness-related activities, THEN Melanie is likely to attend an outdoor fitness event.
      Variables:
      - none: no answer variable is needed because the question asks for a yes/no judgment
      Subgoals:
      - Melanie regularly engages in outdoor exercise.
      - Melanie shows motivation for fitness-related activities.
      Premises:
      - Melanie regularly engages in outdoor exercise.
      - Melanie shows motivation for fitness-related activities.
\end{lstlisting}

\subsubsection{Next-hop decomposition}
\begin{lstlisting}[style=goalmemprompt]
system: |-
  You will be provided with a question about a person and previously generated inference rules.

  Generate one NEW commonsense goal decomposition that is complementary, not redundant, with prior rules.

  Requirements:
  - Do NOT use specific retrieved facts or invented concrete details.
  - Use general human commonsense only.
  - Avoid repeating any prior rule verbatim or semantically.
  - Keep premises/subgoals abstract and checkable from future conversation evidence.
  - Phrase each premise as the kind of EVENT, ACTION, ATTRIBUTE, or ATTITUDE a participant would naturally mention in a casual personal conversation.
  - DO NOT phrase premises as META-CLAIMS about the conversation itself. FORBIDDEN openings include: "the conversation explicitly states ...", "the conversation gives ...", "the message names ...", "a fact says ...", "the speaker explicitly mentions ...".
  - Prefer concrete event/attribute language over evidentiary framing.
  - Preserve exact entity anchors and make answer variables explicit as (x:type), (y:type), etc.

  Output format (exactly):
  Goal: <one goal for the question, using typed variables where needed>
  Rule: IF <subgoal 1> AND <subgoal 2> AND ... THEN <goal is satisfied>
  Variables:
  - (<variable:type>): <what this variable must be substituted with, or "none" if no answer variable is needed>
  Subgoals:
  - <atomic subgoal 1>
  - <atomic subgoal 2>
  Premises:
  - <same text as subgoal 1, repeated for backward compatibility with retrieval code>
  - <same text as subgoal 2, repeated for backward compatibility with retrieval code>
  Difference From Prior Decompositions: <one concise sentence explaining what is new>

input_template_commonsense: |-
  Question: {{QUESTION}}
  {% if BACKGROUND_SUMMARY %}
  Background (per-speaker conversation summary):
  {{BACKGROUND_SUMMARY}}
  Note on the background: this is OPTIONAL context to help you ground the
  speakers — it is NOT exhaustive and NOT a fact source. Do NOT restrict your
  rule to only what appears here, do NOT cite the summary as evidence, and do
  NOT treat absence from the summary as absence from the conversation. General
  commonsense reasoning should still be the primary driver of the rule; use the
  background only as light supporting context.
  {% endif %}
  {% if ASSUMED_KNOWN_INFO %}
  Known Info (assume true): {{ASSUMED_KNOWN_INFO|join('\n')}}
  {% endif %}
  Existing rules/decompositions that should NOT be repeated:
  {{AXIOMS|join('\n')}}
  New goal decomposition:

few-shot:
  - input:
      QUESTION: |-
        What career field is Caroline likely to pursue in her education?
      AXIOMS:
        - IF Caroline consistently expresses sustained interest in (x:career_field) AND Caroline takes concrete steps toward training or education in (x:career_field), THEN Caroline is likely to pursue (x:career_field) in her education.
    output: |-
      Goal: (Caroline) is likely to pursue (x:career_field) in her education.
      Rule: IF Caroline repeatedly describes (x:career_field) as meaningful or personally rewarding, THEN evidence of that repeated valuation would be sufficient additional support for inferring Caroline may pursue (x:career_field).
      Variables:
      - (x:career_field): the education or career field Caroline is likely to pursue
      Subgoals:
      - Caroline repeatedly describes (x:career_field) as meaningful or personally rewarding.
      Premises:
      - Caroline repeatedly describes (x:career_field) as meaningful or personally rewarding.
      Difference From Prior Decompositions: The new decomposition uses repeated valuation rather than training steps.
  - input:
      QUESTION: |-
        Would Melanie be likely to attend an outdoor fitness event?
      AXIOMS:
        - IF Melanie regularly engages in outdoor exercise AND Melanie shows motivation for fitness-related activities, THEN Melanie is likely to attend an outdoor fitness event.
    output: |-
      Goal: (Melanie) is likely to attend an outdoor fitness event.
      Rule: IF Melanie voluntarily participates in social or organized fitness activities, THEN evidence of that participation would be sufficient complementary support for judging likely attendance at a future outdoor fitness event.
      Variables:
      - none: no answer variable is needed because the question asks for a yes/no judgment
      Subgoals:
      - Melanie voluntarily participates in social or organized fitness activities.
      Premises:
      - Melanie voluntarily participates in social or organized fitness activities.
      Difference From Prior Decompositions: The new decomposition uses social or organized participation rather than general outdoor exercise and motivation.
\end{lstlisting}

\subsection{Unification}
\begin{lstlisting}[style=goalmemprompt]
system: |-
  You are answering questions about people based on facts extracted from their personal conversations.

  memory: a self-contained sentence describing an event, action, state, or attribute related to the person. For example, "Caroline is attending therapy sessions" or "Melanie enjoys running in the park".

  This prompt implements the VERIFIABLE UNIFICATION stage of Goal-Mem.

  Task:
  - Given the question, optional goal, optional subgoals, general rules, current substitutions/known info, and retrieved facts, decide whether the retrieved facts ground the goal.
  - Compute any candidate variable substitutions theta = {x / e, ...} that are supported by retrieved facts.
  - Generate the final answer only if the goal is fully grounded by type-consistent, non-conflicting, entailed substitutions.
  - If grounding fails, report unresolved subgoals and known grounded information.
  
  If no explicit Subgoals section is provided, treat the premises of the General Rule as the active subgoals. If no explicit Goal is provided, infer the goal from the Question and General Rule.

  ENTITY-ANCHOR CHECK (do this BEFORE selecting facts):
    1. Identify the question's central entities: its subject, the specific object/topic/instrument/place/activity it asks about, and any qualifier ("previous", "first", named person, time window, etc.).
    2. Use ONLY facts that mention those exact entities or facts that can be unified through an explicit variable in a subgoal. Closely related but distinct entities (guitar vs violin; Korean class on Wednesday vs trip to Korea; current role vs previous role; one party's brownies vs another party's cake) are NOT substitutes.
    3. If no fact mentions the question's central entity and no subgoal variable can validly bridge to it, the goal is not grounded. Do not pick a thematically similar fact as a fallback.

  UNIFICATION PROCESS:
    1. Apply any Current Substitution / Known Info to the active subgoals before evaluating new facts.
    2. For each subgoal psi_i and candidate fact m_j, propose substitutions only for explicit variables such as (x:drink) or (z:cafe).
    3. Type consistency: accept x/e only if e is an instance of the variable type or has a type that entails it in context. For example, Kyoto Latte may fill (x:drink); guitar may not fill (x:instrument asked as violin) unless the subgoal variable is typed broadly as instrument and the question does not require violin.
    4. Equality with existing substitutions: if x is already bound in the current substitution, any new binding for x must be the same entity in context. Reject conflicting bindings.
    5. Logical entailment: after applying the candidate substitution, the retrieved fact must entail the grounded subgoal. Topical similarity is not enough.
    6. Simultaneous consistency: perform the check across all active subgoals and facts as a set. Do not let the order of facts decide which conflicting substitution wins.
    7. Conflict handling: if facts ground the same required variable with incompatible values and the conflict cannot be resolved from the facts alone, answer "I don't know".

  ANSWER RULES:
  - Your answer must be based on the provided facts and general rules/subgoals. State the used facts and rules explicitly in your reasoning.
  - Indicate the number of facts and general rules that you used to answer the question.
  - If the question requires a specific value (date, name, number, location, object, etc.) and the value is not grounded by a successful substitution or an explicit fact, answer "I don't know".
  - Partial or hedged answers ("around that time", "probably", "likely", "it seems") are FORBIDDEN unless the question itself asks for likelihood/preference and the likelihood judgment is grounded by the rule.
  - If the current facts and rules are not sufficient to answer the question, Final Answer must be "I don't know".
  - Do NOT make assumptions beyond the provided facts, rules, and current substitutions.
  - Do NOT output Missing Info or Relation Type here. Use Unresolved Subgoals and Known Info only; refinement is a separate prompt.

  OUTPUT FORMAT — you MUST use exactly this structure:
  Unification Status: <satisfied|unsatisfied|conflict>
  Substitution: <{x / entity, y / entity, ...} or {} or "none">
  Grounded Subgoals:
  - <subgoal> <= <supporting fact # or Known Info>; bindings: <bindings or none>
  Unresolved Subgoals:
  - <subgoal> -- <why it is not grounded, or "none">
  Reasoning: <cite the specific facts and rules that support your answer, or explain why the facts are insufficient; include Used N facts and M general rules>
  Final Answer: <the shortest possible phrase directly supported by the facts/substitution, or "I don't know">
  Known Info: <concise semicolon-separated statements already grounded, or "none">

input_template: |-
  Question: {{QUESTION}}
  {% if GOAL %}
  Goal: {{GOAL}}
  {% endif %}
  {% if SUBGOALS %}
  Subgoals:
  {{ SUBGOALS|join('\n') }}
  {% endif %}
  {% if CURRENT_SUBSTITUTION %}
  Current Substitution: {{CURRENT_SUBSTITUTION}}
  {% endif %}
  {% if ASSUMED_KNOWN_INFO %}
  Known Info (assume true for this depth): {{ ASSUMED_KNOWN_INFO|join('\n') }}
  {% endif %}
  {% if HINTS %}
  Hints: {{ HINTS|join('\n') }}
  {% endif %}
  {% if GENERAL_RULE %}
  General Rule: {{GENERAL_RULE}}
  {% endif %}
  {% if FACTS %}
  Facts: {{ FACTS|join('\n') }}
  {% endif %}

few-shot:
  - input:
      QUESTION: |-
        What career field is Caroline likely to pursue in her education?
        Goal: (Caroline) is likely to pursue (x:career_field) in her education.
        Subgoals:
        - Caroline is actively exploring (x:career_field).
        - Caroline has expressed direct interest in (x:career_field).
        - Caroline finds work in (x:career_field) meaningful.
        General Rule: IF Caroline is actively exploring (x:career_field) AND Caroline has expressed direct interest in (x:career_field) AND Caroline finds work in (x:career_field) meaningful, THEN Caroline is likely to pursue (x:career_field) in her education.
        Facts: 1- (Caroline, exploring, counseling or mental health career options) [event point_in_time: 15 July, 2023] [spoken at: 1:51 pm on 15 July, 2023]
        2- (Caroline, finds, counseling and mental health work tough but rewarding) [event point_in_time: 6 July, 2023] [spoken at: 8:18 pm on 6 July, 2023]
        3- (Caroline, interested_in, counseling) [event point_in_time: 3 July, 2023] [spoken at: 1:36 pm on 3 July, 2023]
    output: |-
      Unification Status: satisfied
      Substitution: {x / counseling and mental health}
      Grounded Subgoals:
      - Caroline is actively exploring (x:career_field). <= Fact 1; bindings: x / counseling and mental health
      - Caroline has expressed direct interest in (x:career_field). <= Fact 3; bindings: x / counseling
      - Caroline finds work in (x:career_field) meaningful. <= Fact 2; bindings: x / counseling and mental health
      Unresolved Subgoals:
      - none
      Reasoning: Fact 1 states Caroline is exploring counseling or mental health career options, Fact 3 states she is interested in counseling, and Fact 2 states she finds counseling and mental health work tough but rewarding. These facts jointly satisfy the three subgoals under the consistent substitution x / counseling and mental health. Used 3 facts and 1 general rule.
      Final Answer: counseling and mental health
      Known Info: Caroline is exploring counseling or mental health career options; Caroline is interested in counseling; Caroline finds counseling and mental health work rewarding
  - input:
      QUESTION: |-
        What is the name of Melanie's dog?
        Goal: (Melanie's dog) has name (x:person_or_pet_name).
        Subgoals:
        - Melanie has a dog.
        - Melanie's dog has name (x:person_or_pet_name).
        General Rule: IF Melanie has a dog AND Melanie's dog has name (x:person_or_pet_name), THEN x is the answer.
        Facts: 1- (Melanie, has_pet, dog) [event point_in_time: 5 August, 2023] [spoken at: 11:20 am on 5 August, 2023]
        2- (Melanie, enjoys, walking her dog in the park) [event point_in_time: 5 August, 2023] [spoken at: 11:20 am on 5 August, 2023]
    output: |-
      Unification Status: unsatisfied
      Substitution: {}
      Grounded Subgoals:
      - Melanie has a dog. <= Fact 1; bindings: none
      Unresolved Subgoals:
      - Melanie's dog has name (x:person_or_pet_name). -- no fact states the dog's name
      Reasoning: Facts 1 and 2 confirm that Melanie has a dog and enjoys walking it, but neither fact states the dog's name. The answer variable x is not grounded. Used 1 fact and 1 general rule.
      Final Answer: I don't know
      Known Info: Melanie has a dog; Melanie enjoys walking her dog in the park
\end{lstlisting}

\subsection{Refinement}
\begin{lstlisting}[style=goalmemprompt]
system: |-
  You are refining unresolved subgoals for Goal-Mem's depth loop.

  You will be provided with a question, the original goal, the current rule/subgoals, the current substitution or known grounded information, retrieved facts, and the unresolved subgoals.

  This prompt implements SUBGOAL REFINEMENT VIA BACKWARD CHAINING.

  Objective:
  - Do NOT answer the user question.
  - Do NOT redo unification.
  - For each unresolved subgoal, generate a more basic requisite antecedent subgoal whose grounding would help satisfy the unresolved subgoal.
  - Produce retrieval-ready Missing Info and Relation Type outputs for the next memory retrieval hop.


  Refinement principles:
  1. Target the exact unresolved subgoal. Preserve all central entities and qualifiers from the question and from the unresolved subgoal.
  2. Do not merely paraphrase the unresolved subgoal. Generate an antecedent that would make the unresolved part checkable.
     Example: unresolved "(x:drink) served in (z:cafe visited last week)" can refine to "Alice visited (z:cafe) last week" if z is unknown.
  3. Keep unresolved variables explicit as (x:type), (y:type), etc. Reuse the same variable names when the refined subgoal is intended to ground the same variable.
  4. Respect existing substitutions. If x is already bound, use the bound entity unless the unification trace says the binding is conflicted.
  5. Do not invent constants. Only use constants that appear in the question, Known Info, Current Substitution, or retrieved facts.
  6. Avoid repeated refinement. If Previously Retrieved Missing Info or Previously Refined Subgoals are provided, the new refinement MUST take a different angle.
  7. If no useful new refinement is possible, set Refinement Status to stop and explain why.

  Relation Type selection:
  - temporal: the missing evidence is about when something happened or the sequence of events.
  - causal: the missing evidence is about why something happened, what caused it, or what resulted from it.
  - semantic: the missing evidence is about the same specific topic, activity, object, name, preference, property, or event.
  - spatial: the missing evidence is about where something happened, a place identity, or what else happened at the same location.

  OUTPUT FORMAT — you MUST use exactly this structure:
  Refinement Status: <refine|stop>
  Missing Info: <specific information needed for the next retrieval hop>
  Relation Type: <temporal|causal|semantic|spatial>
  Refined Subgoals:
  - <atomic subgoal to retrieve next, or "none">
  Retrieval Queries:
  - <natural-language query corresponding to refined subgoal 1, or "none">
  Rationale: <one concise sentence explaining why the refined subgoal is necessary>

input_template: |-
  Question: {{QUESTION}}
  {% if GOAL %}
  Goal: {{GOAL}}
  {% endif %}
  {% if GENERAL_RULE %}
  General Rule: {{GENERAL_RULE}}
  {% endif %}
  {% if SUBGOALS %}
  Current Subgoals:
  {{SUBGOALS|join('\n')}}
  {% endif %}
  {% if CURRENT_SUBSTITUTION %}
  Current Substitution: {{CURRENT_SUBSTITUTION}}
  {% endif %}
  {% if KNOWN_INFO %}
  Known Info:
  {{KNOWN_INFO|join('\n')}}
  {% endif %}
  {% if UNIFICATION_TRACE %}
  Unification Trace:
  {{UNIFICATION_TRACE}}
  {% endif %}
  {% if UNRESOLVED_SUBGOALS %}
  Unresolved Subgoals:
  {{UNRESOLVED_SUBGOALS|join('\n')}}
  {% endif %}
  {% if PREVIOUSLY_RETRIEVED_MISSING_INFO %}
  Previously Retrieved Missing Info (do NOT repeat or paraphrase):
  {{PREVIOUSLY_RETRIEVED_MISSING_INFO|join('\n')}}
  {% endif %}
  {% if PREVIOUSLY_REFINED_SUBGOALS %}
  Previously Refined Subgoals (do NOT repeat or paraphrase):
  {{PREVIOUSLY_REFINED_SUBGOALS|join('\n')}}
  {% endif %}
  {% if FACTS %}
  Retrieved Facts:
  {{FACTS|join('\n')}}
  {% endif %}
  Refinement:

few-shot:
  - input:
      QUESTION: |-
        Alice: I like the cafe I went to last week, what drink should I try this time?
      GOAL: |-
        Recommend (x:drink) to (Alice) from the menu of the cafe she went to last week.
      GENERAL_RULE: |-
        IF Alice likes (y:flavor) AND (x:drink) contains (y:flavor) AND (x:drink) is served in (z:cafe visited last week), THEN recommend (x:drink) to Alice.
      SUBGOALS:
        - Alice likes (y:flavor).
        - (x:drink) contains (y:flavor).
        - (x:drink) is served in (z:cafe visited last week).
      CURRENT_SUBSTITUTION: "{x / Kyoto Latte, y / Matcha Powder}"
      KNOWN_INFO:
        - Alice likes matcha.
        - Kyoto Latte contains Matcha Powder.
      UNRESOLVED_SUBGOALS:
        - (x:drink) is served in (z:cafe visited last week). -- z is not grounded; no fact identifies the cafe Alice visited last week
      FACTS:
        - "1- The Momoco cafe advertised a Kyoto Latte as its seasonal drink."
    output: |-
      Refinement Status: refine
      Missing Info: Which cafe Alice visited last week, so z can be grounded for the served-in subgoal.
      Relation Type: spatial
      Refined Subgoals:
      - Alice visited (z:cafe) last week.
      Retrieval Queries:
      - Alice visited cafe last week which cafe
      Rationale: The unresolved served-in subgoal cannot be checked until the variable z:cafe visited last week is grounded.
  - input:
      QUESTION: |-
        What is the name of Melanie's dog?
      GOAL: |-
        (Melanie's dog) has name (x:person_or_pet_name).
      GENERAL_RULE: |-
        IF Melanie has a dog AND Melanie's dog has name (x:person_or_pet_name), THEN x is the answer.
      CURRENT_SUBSTITUTION: "{}"
      KNOWN_INFO:
        - Melanie has a dog.
      UNRESOLVED_SUBGOALS:
        - Melanie's dog has name (x:person_or_pet_name). -- no fact states the dog's name
      FACTS:
        - "1- (Melanie, has_pet, dog) [event point_in_time: 5 August, 2023] [spoken at: 11:20 am on 5 August, 2023]"
        - "2- (Melanie, enjoys, walking her dog in the park) [event point_in_time: 5 August, 2023] [spoken at: 11:20 am on 5 August, 2023]"
    output: |-
      Refinement Status: refine
      Missing Info: The name Melanie uses for her dog.
      Relation Type: semantic
      Refined Subgoals:
      - Melanie mentions her dog by name as (x:person_or_pet_name).
      Retrieval Queries:
      - Melanie dog name mentions her dog by name
      Rationale: The dog is established, but the answer variable x remains unbound until a memory names the dog.
\end{lstlisting}
\end{document}